%% file: main.tex
\documentclass[runningheads]{llncs}

\usepackage[T1]{fontenc}
\usepackage{amsmath}
\usepackage{amssymb}
\usepackage{cite}
\usepackage{xspace}
\usepackage[hyphens]{url}

\usepackage{graphicx}
\usepackage{booktabs}
\usepackage{enumitem}
\usepackage{multirow}
\usepackage{algorithm}
\usepackage{algpseudocode}
\usepackage{wrapfig}  
\usepackage{adjustbox}  
\usepackage{xcolor}

\usepackage[accsupp]{axessibility}  

\makeatletter
\DeclareRobustCommand\onedot{\futurelet\@let@token\@onedot}
\def\@onedot{\ifx\@let@token.\else.\null\fi\xspace}
\def\eg{\emph{e.g}\onedot}

\def\ie{\emph{i.e}\onedot}

\makeatother

\usepackage{hyperref}
\hypersetup{hidelinks}
\definecolor{projectpink}{HTML}{D63384}
\newcommand{\projectpage}{\href{https://tenplusgood.github.io/a-harness-page/}{\textcolor{projectpink}{Project Page}}}

\begin{document}

\title{Affordance Agent Harness: Verification-Gated Skill Orchestration} 

\titlerunning{Affordance Agent Harness}

\author{Haojian Huang\inst{1,2}\textsuperscript{*} \and
Jiahao Shi\inst{2,3}\textsuperscript{*} \and
Yinchuan Li\inst{1,2} \and
Yingcong Chen\inst{1,2}\textsuperscript{\dag}}

\authorrunning{Preprint}

\institute{\textsuperscript{1}HKUST(GZ) \quad
\textsuperscript{2}Knowin AI \quad
\textsuperscript{3}Harbin Engineering University}

\maketitle
\begingroup
\renewcommand\thefootnote{\fnsymbol{footnote}}
\footnotetext[1]{Equal contribution.}
\footnotetext[4]{Corresponding author.}
\endgroup
\vspace{-0.6em}

\begin{abstract}
Affordance grounding requires identifying \emph{where} and \emph{how} an agent should interact in open-world scenes, where actionable regions are often small, occluded, reflective, and visually ambiguous. Recent systems therefore combine multiple skills (\eg, detection, segmentation, interaction-imagination), yet most orchestrate them with fixed pipelines that are poorly matched to per-instance difficulty, offer limited targeted recovery from intermediate errors, and fail to reuse experience from recurring objects. These failures expose a systems problem: test-time grounding must acquire the right evidence, decide whether that evidence is reliable enough to commit, and do so under bounded inference cost without access to labels. We propose \textbf{Affordance Agent Harness}, a closed-loop runtime that unifies heterogeneous skills with an evidence store and cost control, retrieves episodic memories to provide priors for recurring categories, and employs a Router to adaptively select and parameterize skills. An affordance-specific Verifier then \emph{gates commitments} using self-consistency, cross-scale stability, and evidence sufficiency, triggering targeted retries before a final judge fuses accumulated evidence and trajectories into the prediction. Experiments on multiple affordance benchmarks and difficulty-controlled subsets show a stronger accuracy--cost Pareto frontier than fixed-pipeline baselines, improving grounding quality while reducing average skill calls and latency. \projectpage.
\end{abstract}

\section{Introduction}
\label{sec:intro}
Affordance grounding aims to identify \emph{where} and \emph{how} an agent should interact in open-world visual scenes. In everyday environments, actionable regions---handles, buttons, rims, and graspable edges---are often small, partially occluded, reflective, or visually ambiguous~\cite{wu2024afforddp,yu2025seqafford,zhou2025r1}. This demands fine-grained spatial localization, high-level semantic reasoning, and robust recovery from intermediate perception errors~\cite{wang2025affordance}. Despite recent progress of large multimodal models (LMMs) in recognition and language-grounded reasoning~\cite{guo2025deepseekr1,gpt4,qwen3.5,yang2025qwen3}, affordance grounding remains brittle when decisions must be geometry-aware and evidence must be aggregated coherently across heterogeneous cues and tools.

\begin{wrapfigure}{r}{0.5\textwidth}
    \centering
    \vspace{-0.8em}
    \includegraphics[width=0.50\textwidth]{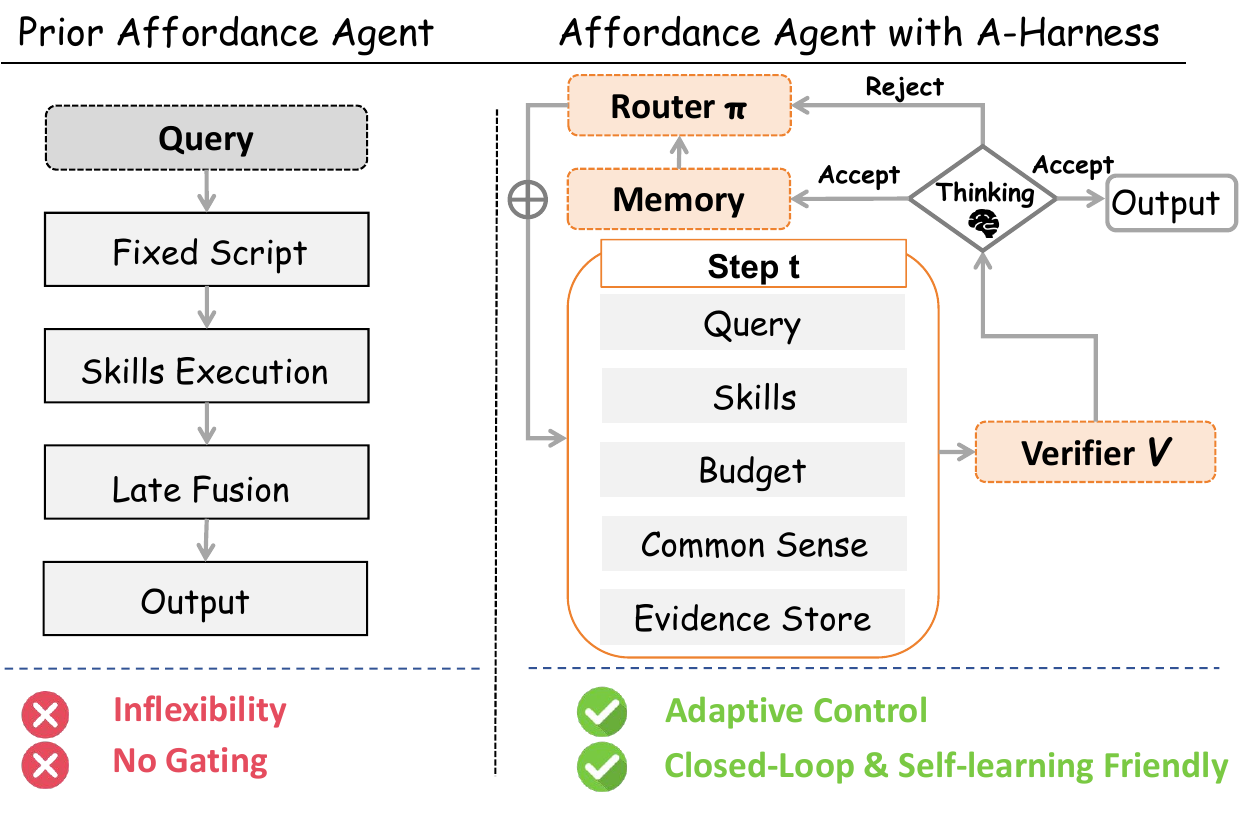}
    \caption{Comparison between a prior affordance agent with a fixed reasoning graph and our A-Harness–enabled agent. While prior systems execute skills along a predefined script with late fusion and no commitment gating, A-Harness introduces a context-aware, budgeted closed-loop runtime with adaptive routing, verification-driven retries, and persistent memory for reusable experience.}
    \label{fig:comp}
    \vspace{-0.8em}
\end{wrapfigure}

Prior work has pursued two main directions. One line fine-tunes LMMs for affordance estimation, but scarce affordance annotations often lead to substantial degradation on unseen open-world instances. A more promising line adopts an \emph{agent paradigm}, leveraging increasingly capable generative and editing models (\eg \cite{wu2025qwenimagetechnicalreport,zheng2024videogen}) to enrich the reasoning context at inference time~\cite{wang2025affordance,zhang2025a4agent}. Existing affordance agents, however, still inherit three failure modes from fixed perception pipelines. First, they execute skills in a predetermined order: easy cases are over-processed, while hard cases such as small targets, occlusion, or OOD categories do not receive targeted additional evidence. Second, they lack closed-loop correction: when detector boxes, segmentation masks, or interaction points disagree across tools and scales, systems typically fuse results late instead of identifying the failure source and re-invoking the responsible skill, allowing silent errors to propagate~\cite{zhao2026star,huang2026find}. Third, they lack persistent experience: recurring objects and categories are re-solved from scratch, amortizing neither successful tool chains nor object-level priors (see Fig.~\ref{fig:comp}).

Together, these limitations recast the task as budgeted evidence seeking rather than single-pass prediction. Without ground-truth labels at test time, ``verification'' cannot certify absolute correctness; instead, it must provide \emph{relative} and \emph{actionable} diagnostics that decide whether current evidence is trustworthy enough to commit and, if not, what evidence is missing. This view is aligned with evidence-centric fusion and uncertainty-guided reasoning, where reliability is inferred from agreement, discounting, and targeted re-querying rather than a single raw prediction~\cite{huang2023belief,liu2024adaptive,huang2024evidential,chen2024uncertainty}. We therefore rely on three signal families: (\textit{i}) \textbf{cross-tool consistency}, measuring agreement between detection, segmentation, and other cues within the same region; (\textit{ii}) \textbf{cross-scale stability}, measuring whether actionable regions or interaction points persist under zoom-in/out and re-cropping; and (\textit{iii}) \textbf{evidence sufficiency}, measuring whether the evidence store contains enough supported and traceable observations for the decision. These signals do not merely score outputs; they decide whether to commit or to launch a targeted retry, such as zooming into a disputed region, re-running detection with class constraints, or invoking external knowledge retrieval when semantics are ambiguous.

A-Harness implements this process as a closed-loop execution framework that separates a reproducible runtime from a decision policy. The runtime unifies heterogeneous skills (\eg, detection, segmentation, zoom-in, web search, and interaction imagination) under a shared interface, maintains an evidence store with provenance, enforces budgets, and records reasoning trajectories. Given an input, the system retrieves relevant \emph{episodic memories} to provide \emph{actionable priors}---successful tool chains and parameters for similar objects and tasks---which enables fast paths for recurring categories. A \emph{Router} then selects the next skill and its parameters (\eg, region of interest, prompts, search queries) conditioned on the current evidence state and remaining budget. After each execution, an affordance-specific \emph{Verifier} applies the diagnostics above to localize likely failure modes and either accept the evidence or request a corrective action. When the confidence criteria are satisfied, or the budget is exhausted, a final policy fuses structured evidence, trajectories, and retrieved memory into an affordance prediction with traceable provenance. Verified episodes are written back to memory, so the system can reduce exploration cost on similar future cases.

Our contributions are threefold:
\begin{itemize}
    \item We replace fixed pipelines with a per-instance routing policy over heterogeneous skills and their parameters for affordance grounding.
    \item We introduce relative, actionable verification signals (consistency, stability, sufficiency) to gate commitments and trigger targeted retries, and an episodic memory that amortizes successful tool chains for recurring objects and tasks.
    \item  On multiple benchmarks and difficulty-controlled subsets, we achieve higher grounding accuracy with fewer skill calls and lower latency than fixed-pipeline baselines, supported by ablations isolating verification and memory.
\end{itemize}

\section{Related Work}
\label{sec:related_work}
Affordance grounding has shifted from category-specific, fully supervised region predictors to \emph{open-world} inference, where actionable regions are tiny, occluded, reflective, and distribution-shifted. Early work established end-to-end affordance detection/segmentation with curated annotations and part-level supervision on images and 3D shapes~\cite{do2017affordancenet,deng20213d,mo2019partnet,xu2022partafford}. To mitigate annotation bottlenecks and improve generalization, later studies explored weakly/annotation-free learning and few-shot transfer for articulated objects under occlusions, as well as leveraging human behaviors and demonstrations as scalable supervision for functional contact cues~\cite{mo2022o2o,ning2023where2explore,wu2023learning,bahl2023affordances,luo2023learning,chen2023affordance}. In parallel, the field expanded to richer 3D and long-horizon formulations that require intent-conditioned reasoning over \emph{multiple} candidate interaction sites and sequential affordances, together with open-vocabulary settings that connect language and geometry for unseen affordance concepts~\cite{yang2023grounding,delitzas2024scenefun3d,yu2025seqafford,nguyen2023open,van2024open,shao2024great,chu20253d}. A complementary line treats affordances as transferable \emph{manipulation priors} bridging perception and control, improving policy generalization yet remaining sensitive to noisy grounding evidence in the wild~\cite{nasiriany2024rt,wu2024afforddp}. More recent efforts increasingly exploit pretrained vision-language/foundation models via language-conditioned grounding, distillation, and one-shot adaptation, pushing toward open-world robustness but repeatedly surfacing a key bottleneck: heterogeneous intermediate evidence can be inconsistent, and without test-time labels systems lack an operational notion of \emph{verification} to decide when to commit versus when (and where) to gather more evidence~\cite{qian2024affordancellm,tong2024oval,li2024one,gao2024learning,lu2024geal,wang2025dag,wu2025ragnet,wang2025affordance}. Meanwhile, promptable segmentation and open-vocabulary localization, together with instruction-tuned MLLMs and region-level grounding, make it practical to compose detection/segmentation/zoom-in skills at inference time rather than retraining bespoke affordance models~\cite{kirillov2023segment,ravi2024sam,liu2024grounding,minderer2023scaling,liu2024visual,zhang2023gpt4roi,you2023ferret,lai2024lisa,carion2025sam}. The ``think-with-pixels'' trend further strengthens this direction by treating pixel-space intermediate steps as part of reasoning, suggesting that tool-mediated, dynamically refreshed visual context can be a first-class driver of robust open-world grounding~\cite{ren2024pixellm,pixelreasoner}.
At the agent-system level, tool-using and embodied frameworks highlight iterative self-correction, verification, and experience reuse as mechanisms for reliability and efficiency~\cite{yao2022react,ahn2022saycan,huang2023voxposer,driess2023palme,schick2023toolformer,xu2023rewoo,shinn2023reflexion,hallinan2023selfrefine}.
In embodied settings in particular, recent vision-language-action models and embodied MLLMs further strengthen this trend by grounding language in 3D perception and manipulation, enabling agents to plan and act with richer, task-conditioned visual/multimodal context~\cite{li2024manipllm,zhen20243d,hong20233d,qi2024shapellm,xu2023pointllm,he2024segpoint,Dwivedi_2025_CVPR,song2025maniplvm,qwen25vl,internvl3}.
Orthogonal evidence-centric research studies Dempster--Shafer fusion, evidential multi-view clustering/classification, and reliability discounting for uncertain observations~\cite{huang2023belief,liu2023adaptive,liu2024adaptive,huang2024evidential,huang2025trusted}; related multimodal reasoning work further uses uncertainty-guided self-questioning, structured spatial correction, and context repair to reduce cascading errors~\cite{chen2024uncertainty,zhao2026star,huang2026find}.
However, existing affordance agents still largely behave as fixed or weakly adaptive pipelines~\cite{zhang2025a4agent}: conflicts (\eg, box--mask mismatch) are handled by late fusion rather than diagnosis-driven re-invocation; tool usage is rarely optimized as a budgeted policy; and experience is seldom amortized as \emph{actionable priors} (skill chains and parameters) for recurring categories.
These gaps motivate our \textbf{Affordance Agent Harness}, which makes budget-aware routing, affordance-specific verification that \emph{gates commitments} and triggers targeted retries, and episodic memory over executable experience first-class components of a closed-loop runtime, improving the accuracy--cost trade-off in open-world affordance grounding.

\vspace{-0.4em}
\section{Method}
\label{sec:method}
\vspace{-0.3em}
\subsection{Problem Setup}
\vspace{-0.3em}
Knowing \emph{where} to interact with an object---and \emph{how}---is a prerequisite for any embodied agent that must act in the physical world, yet affordance remains one of the least standardized prediction targets in vision.
Given an observation $x$ (an image or a set of views) and an optional instruction $q$, affordance grounding predicts an \emph{actionable region} $A$ describing where and how to interact.
We focus on pixel-level outputs and represent the affordance as a dense map $A\in[0,1]^{H\times W}$ in the image coordinate frame (binary or soft masks), while the framework naturally extends to keypoints or contact regions.
\begin{figure}[!t]
    \centering
    \includegraphics[width=\textwidth]{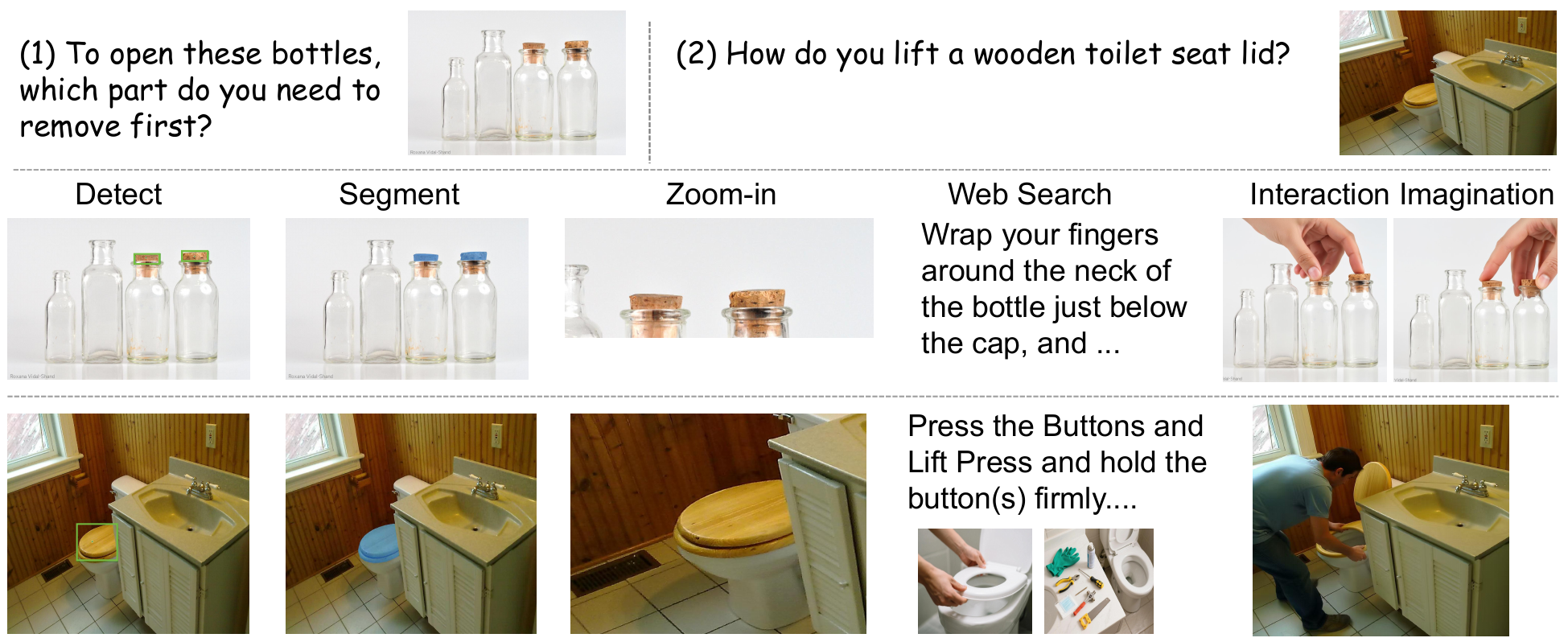}
    \vspace{-1em}
    \caption{Illustration of heterogeneous skills that generate complementary visual and semantic evidence. Web search can retrieve both textual guidance and paired images when available (\ie case(2)), enriching the visual context for affordance reasoning.}
    \label{fig:skill_illustration}
    \vspace{-0.6em}
\end{figure}
We assume access to a set of general-purpose skills $\mathcal{S}=\{s_1,\dots,s_K\}$---detection, segmentation, zoom-in, web search, and interaction imagination (Dreamer)---that can be invoked at test time (see Fig.~\ref{fig:skill_illustration}).
Each invocation incurs cost; we adopt a primary budget $B$ measured by the number of skill calls, and track latency and monetary cost as secondary.

\vspace{-1em}
\subsection{Overview}
\vspace{-0.3em}
\label{sec:overview}
No single skill reliably localizes affordances in the open world: detectors can miss small functional parts, segmenters can refine the wrong region, and external knowledge can be semantically useful but spatially weak. A-Harness addresses this mismatch by acquiring, cross-checking, and fusing complementary evidence under a budget before committing to a prediction.

\begin{figure}[!t]
    \centering
    \includegraphics[width=\textwidth]{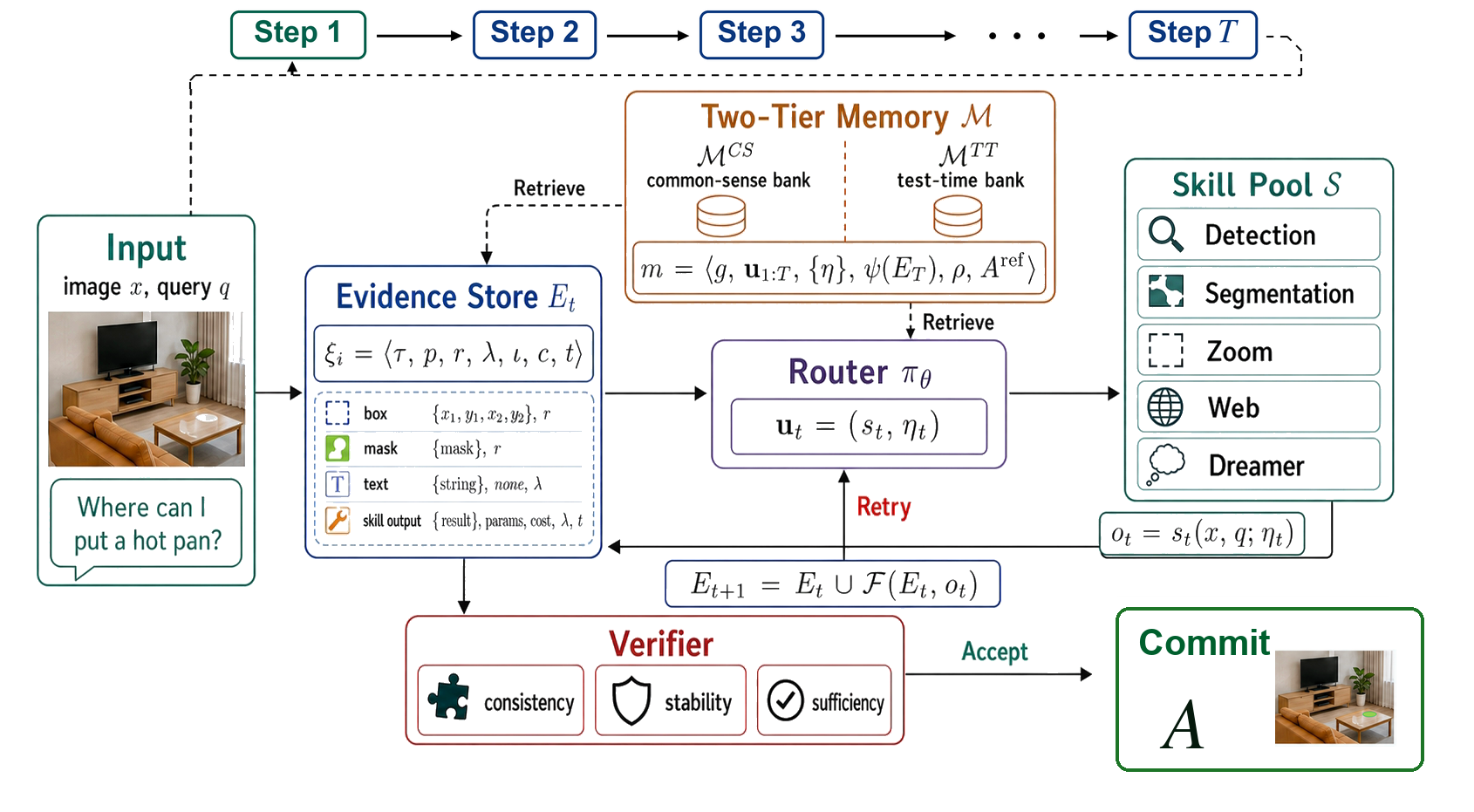}
    \vspace{-0.5em}
    \caption{Overview of the A-Harness framework, illustrating iterative decision-making. The Verifier dynamically assesses evidence, guiding the Router to either re-plan or output results, while storing the trajectory in memory. The skill outcome $o_t$ is stored in the evidence store and combined with existing evidence to support the next step.}
    \label{fig:framework}
    \vspace{-0.6em}
\end{figure}

As shown in Fig.~\ref{fig:framework}, the system includes four components: (i) an \textbf{evidence store}~$E_t$ that accumulates heterogeneous skill outputs, (ii) a \textbf{two-tier memory}~$\mathcal{M}$ providing reusable priors, (iii) a \textbf{Router}~$\pi_\theta$ for skill selection, and (iv) a \textbf{Verifier} for gating commitment and triggering retries. At step~$t$, the Router selects an action $u_t=(s_t,\eta_t)$, where $s_t\in\mathcal{S}$ is the skill and $\eta_t$ its parameters. Evidence is updated as:
\begin{equation}
u_t \sim \pi_\theta(\cdot \mid z_t), \quad o_t = s_t(x,q;\eta_t), \quad E_{t+1} = E_t \cup \mathcal{F}(E_t,o_t),
\end{equation}
where $u_t$ is the action selected by the Router at time $t$ based on state $z_t$, $s_t$ is the chosen skill with parameters $\eta_t$, $o_t$ is the output, and $\mathcal{F}(E_t,o_t)$ is a parsing function that extracts evidence-relevant observations from $o_t$ and wraps them into typed evidence records (Eq.~\ref{eq:eitem_final}). The loop ends when the Verifier accepts or the budget is exhausted, outputting $A$.

\vspace{-1.2em}
\subsection{Details of A-Harness}
\vspace{-0.4em}
\label{subsec:details}
\paragraph{Evidence Store with Provenance.}
\label{para:evidence}
As skills are invoked, the system accumulates a heterogeneous mixture of outputs---boxes from a detector, masks from a segmenter, textual cues from web search---that must be organized before they can be cross-checked or fused, echoing the need for provenance-aware multi-source evidence fusion~\cite{huang2023belief}.
The evidence store $E_t$ addresses this by tagging every intermediate result with its provenance.
Each item is a record
\begin{equation}
\xi=\langle \tau,\, p,\, r,\, \lambda,\, \iota,\, c,\, t \rangle ,
\label{eq:eitem_final}
\end{equation}
where $\tau$ is the evidence type (\eg, box/mask/text), $p$ is the payload (\ie the actual value), $r$ is the ROI,
$\lambda$ is the scale/zoom level, $\iota\in\{1,\dots,K\}$ is the \emph{producer id} identifying which skill generated this item,
$c$ is the incurred cost, and $t$ is the step index.
The producer id~$\iota$ is key: it enables cross-skill agreement checks and tells the system \emph{which} skill to rerun when conflicts arise.
\vspace{-0.4em}
\paragraph{Two-Tier Memory with Feature Retrieval.}
\label{para:memory}
Open-world affordance grounding exhibits a distinctive pattern: most queries involve familiar objects whose affordances follow stable part-level regularities, yet the tail is long and unpredictable---unusual designs, heavy occlusion, or novel tool--object pairings can defeat any fixed prior.
A-Harness amortizes experience via a two-tier memory that separates \emph{stable priors} from \emph{online adaptation}: a \textbf{common-sense bank}~$\mathcal{M}^{\textsc{cs}}$ and a \textbf{test-time episodic bank}~$\mathcal{M}^{\textsc{tt}}$.

$\mathcal{M}^{\textsc{cs}}$ stores prototypical affordance priors for frequent objects and parts collected offline, each paired with an annotated reference mask.
At test time no ground truth is available; instead, $\mathcal{M}^{\textsc{tt}}$ accumulates \emph{verifier-accepted} episodes---successful action sequences and their evidence summaries---so that the agent can adapt to novel challenges on the fly.
Each memory entry stores:
\begin{equation}
m=\langle g,\mathbf{u}_{1:T},\{\eta\},\psi(E_T),\rho,A^{\mathrm{ref}}\rangle .
\label{eq:mem_final}
\end{equation}
where $g$ is a retrieval embedding, $\mathbf{u}_{1:T}$ a successful action sequence, $\{\eta\}$ effective parameter ranges, $\psi(E_T)$ a verified evidence summary, $\rho$ an outcome score (Eq.~\ref{eq:v_final}), and $A^{\mathrm{ref}}$ an optional reference mask (present for~$\mathcal{M}^{\textsc{cs}}$, typically absent for~$\mathcal{M}^{\textsc{tt}}$).

For retrieval we combine DINO-v3~\cite{DINOv3} and SigLIP~\cite{SigLIP} features into a joint embedding:
\begin{equation}
g(x,q) \triangleq \mathrm{norm}\!\left(\, \mathbf{f}_{\textsc{d}}(x)\ \Vert\ \mathbf{f}_{\textsc{c}}(x,q)\,\right),
\qquad
\mathrm{sim}(g,g') = g^\top g' ,
\label{eq:retrieval_final}
\end{equation}
and retrieve the top-$N$ entries by cosine similarity to seed the Router with priors. Episodes are written back to $\mathcal{M}^{\textsc{tt}}$ only when the Verifier accepts.

\textbf{Metabolism.}
Both banks have fixed capacities $|\mathcal{M}^{\textsc{cs}}|\le C_{\textsc{cs}}$ and $|\mathcal{M}^{\textsc{tt}}|\le C_{\textsc{tt}}$.
When $\mathcal{M}^{\textsc{tt}}$ exceeds capacity, the oldest entry is evicted and compressed into a compact \emph{experience capsule} (summarized $\psi(\cdot)$ and robust parameter ranges) that is stored back, preserving long-term utility within bounded storage.

\paragraph{Router: Budget-Aware Skill Selection and Parameterization.}
\label{para:router}
In principle the agent could invoke every skill on every input; in practice each call costs time and money, and many would be redundant.
The Router~$\pi_\theta$ maps the current state~$z_t$ to the next action $u_t=(s_t,\eta_t)$, choosing the skill most likely to resolve the current uncertainty per unit cost.
A-Harness is agnostic to how $\pi_\theta$ is realized---heuristic or learned---and makes budget trade-offs explicit by ranking candidates via an estimated benefit--cost ratio:
\begin{equation}
u_t = \arg\max_{u\in\mathcal{U}(z_t)} \frac{\widehat{\Delta v}(u;z_t)}{\widehat{c}(u)} ,
\label{eq:utility_final}
\end{equation}
where $\mathcal{U}(z_t)$ is the feasible action set under the current evidence and remaining budget,
$\widehat{\Delta v}(u;z_t)$ estimates the expected increase in verifier score,
and $\widehat{c}(u)$ estimates cost (typically $1$ per call, optionally latency-weighted).
In practice, whenever commitment is denied, the Verifier emits a diagnostic indicating the dominant gap (e.g.\ low consistency or insufficient evidence) together with a suggested remedial action; the Router consults this proposal and selects the $u$ that best addresses the flagged gap under the remaining budget. The full instantiation of $\widehat{\Delta v}$, including the verifier-gap heuristic and LLM fallback, is detailed in Appendix~\ref{app:delta_v}.

\paragraph{Verifier: Relative Diagnostics and Commitment Gating.}
\label{para:verifier}
The central challenge of test-time verification is the absence of ground truth.
We sidestep this by designing \emph{relative} diagnostics---signals that correlate with correctness by measuring agreement, stability, and sufficiency of the accumulated evidence rather than comparing against a label.
The design is motivated by a recurring empirical pattern: failures in open-world grounding are typically preceded by (i)~disagreement across tools, (ii)~prediction drift under re-cropping or zooming, or (iii)~premature commitment from sparse evidence, consistent with evidential discounting and trusted multi-view dynamics under partial or noisy observations~\cite{liu2023adaptive,huang2024evidential,huang2025trusted}.

\textbf{Cross-skill consistency ($\omega_t$).}
Different skills observe the same scene through different inductive biases; when they converge on the same region, the prediction is far more likely to be correct.
Let $\mathcal{B}_t$ be the latest set of boxes and $\mathcal{K}_t$ the latest set of masks within the same ROI, both in a common coordinate frame.
We measure agreement by
\begin{equation}
\omega_t = \max_{b\in\mathcal{B}_t,\; m\in\mathcal{K}_t} \mathrm{IoU}\!\left(\chi(b),\, m\right),
\label{eq:c_final}
\end{equation}
where $\chi(\cdot)$ converts a box to its filled mask. (For point outputs, IoU is replaced by normalized point distance.)

\textbf{Cross-scale stability ($\zeta_t$).}
Hard cases---small targets, heavy occlusion---often produce predictions that ``wander'' when the agent zooms in or shifts the crop.
We enforce stability across scales by comparing predictions at different zoom levels after projecting them back to the original image frame.
Let $\Pi_t$ be a set of cross-scale prediction pairs and $\varphi(\xi)$ the predicted mask of evidence~$\xi$ in the global frame. We quantify drift by
\begin{equation}
\zeta_t = 1-\frac{1}{|\Pi_t|}\sum_{(\xi_i,\xi_j)\in\Pi_t} d\!\left(\varphi(\xi_i),\,\varphi(\xi_j)\right),
\label{eq:s_final}
\end{equation}
where $d(\cdot,\cdot)$ is IoU-distance for masks (or normalized Euclidean distance for points).

\textbf{Evidence sufficiency ($\mu_t$).}
Even when a single tool output looks plausible, committing too early is risky in the open world.
We estimate whether the current hypothesis is backed by \emph{enough} independent evidence.
Let $h_t$ denote the best mask aggregated so far. We define
\begin{equation}
\mu_t = \sum_{\xi\in E_t} w_{\tau(\xi)}\, \kappa(\xi,h_t),
\label{eq:u_final}
\end{equation}
where $w_{\tau}$ is a type-dependent weight and $\kappa(\xi,h_t)\in\{0,1\}$ is an interpretable support indicator---whether $\xi$ overlaps the hypothesis ROI, whether at least one high-confidence localizer corroborates it, and whether no unresolved conflict remains.

\textbf{Choosing $w_{\tau}$.} We let the agent decide $w_{\tau}$ at test time based on evidence quality and its marginal contribution toward a verifiable state, which varies across skills. Details are provided in Appendix~\ref{app:w_tau}.

\textbf{Commit rule.}
At step $t$ we compute three diagnostics: consistency $\omega_t$, stability $\zeta_t$, and sufficiency $\mu_t$ (all scaled to $[0,1]$), and aggregate them as
\begin{equation}
v_t=\alpha\,\omega_t+\beta\,\zeta_t+\gamma\,\sigma(\mu_t),
\label{eq:v_final}
\end{equation}
with weights $\alpha,\beta,\gamma\ge 0$ and a squashing function $\sigma(\cdot)$.
The system \emph{commits} (terminates evidence collection and outputs via $\Phi$) iff
\vspace{-0.4em}
\begin{equation}
v_t \ge \delta \ \wedge\  \omega_t \ge \underline{\omega},
\label{eq:commit}
\end{equation}
where $\delta$ is a global acceptance threshold and $\underline{\omega}$ is a hard floor preventing commitment under tool disagreement.

\textbf{Targeted retry.}
If not committed, we pick the \emph{dominant deficiency} via its margin-to-threshold:
\vspace{-0.4em}
\begin{equation}
\ell_t=\arg\max_{j\in\{\omega,\zeta,\mu\}}\bigl(\underline{j}-j_t\bigr)_+,
\qquad
u_{prop}\leftarrow \Gamma(\ell_t),
\label{eq:diag_final}
\end{equation}
where $(x)_+=\max(x,0)$, and $\Gamma(\cdot)$ maps each failure mode to a corrective action (\eg, $\omega$-failure $\to$ ROI refinement and re-segmentation; $\zeta$-failure $\to$ zoom-in or re-crop; $\mu$-failure $\to$ invoke complementary knowledge/imagination). The Router then selects the final $u$ informed by this proposal, following the same principle as context repair: identify the missing or corrupted evidence and patch it with the smallest targeted intervention~\cite{huang2026find}.
\paragraph{Final Fusion under Budget.}
\label{sec:judge}
Once the loop terminates---either by acceptance or budget exhaustion---we produce the affordance map by fusing the structured evidence and memory priors:
\begin{equation}
A = \Phi\!\left(x,\,q,\,E_T,\,\mathbf{u}_{1:T},\,\mathcal{M}(x,q)\right).
\label{eq:phi_final}
\end{equation}
$\Phi$ can be a deterministic policy (\eg, selecting the most scale-stable mask or averaging consistent hypotheses) or a lightweight model/MLLM that reasons over \emph{structured} evidence rather than re-solving the problem from raw pixels.

In our implementation, $\Phi$ operates in two stages.
First, the accumulated evidence (\ie reference masks, web-retrieved images, textual cues, and cross-scale predictions) is assembled into a rich visual-semantic context that allows the detection skill to localize an accurate keypoint on the actionable region. Then a segmentation foundation model (\ie SAM~2) takes the keypoint as a prompt and produces the final mask~$A$.
Algorithm~\ref{alg:aharness} summarizes A-Harness inference.

\begin{algorithm}[t]
\caption{A-Harness Inference (Budgeted Closed-Loop Execution)}
\label{alg:aharness}
\begin{algorithmic}[1]
\Require observation $x$, instruction $q$, skills $\mathcal{S}$, budget $B$, memory $\mathcal{M}=\{\mathcal{M}^{\textsc{cs}},\mathcal{M}^{\textsc{tt}}\}$
\State $E \gets \emptyset$;\quad $b \gets B$;\quad $u_{\mathrm{prop}} \gets \emptyset$ \Comment{no Verifier proposal initially}
\State Compute retrieval embedding $g(x,q)$ (Eq.~\ref{eq:retrieval_final}) and retrieve top-$N$ entries from $\mathcal{M}$
\For{$t=1,2,\dots$}
    \State $z \gets \langle x,q,E,\mathcal{M}(x,q),b\rangle$
    \State Choose $u=(s,\eta)$ via Router (Eq.~\ref{eq:utility_final}), informed by $u_{\mathrm{prop}}$
    \State Execute $o_t \gets s(x,q;\eta_t)$ and update $E_{t+1} = E_t \cup \mathcal{F}(E_t,o_t)$
    \State $b \gets b - \widehat{c}(u)$
    \State Compute $\omega_t,\zeta_t,\mu_t$ and $v_t$ (Eqs.~\ref{eq:c_final}--\ref{eq:v_final})
    \If{$(v_t\ge\delta \ \wedge\  \omega_t\ge\underline{\omega})$ \textbf{or} $b\le 0$}
        \State \textbf{break}
    \EndIf
    \State Diagnose $\ell_t$; Verifier proposes $u_{\mathrm{prop}} \leftarrow \Gamma(\ell_t)$ (Eq.~\ref{eq:diag_final})
\EndFor
\State Output affordance map $A \gets \Phi(x,q,E,\mathbf{u}_{1:t},\mathcal{M}(x,q))$ (Eq.~\ref{eq:phi_final})
\If{$v_t\ge\delta$}
    \State Write back verified episode to $\mathcal{M}^{\textsc{tt}}$
\EndIf \\
\Return $A$
\end{algorithmic}
\end{algorithm}
\vspace{-0.5em}
\subsection{Discussion}
\vspace{-0.5em}
\label{subsec:Discussion}
A natural question is whether such a runtime is necessary when end-to-end models can map pixels to affordances directly.
We view the two paradigms as complementary: learned models are efficient and strong in-distribution, yet often degrade on the open-world tail (\eg, novel objects, rare viewpoints, or unseen tool--object pairings) that dominates deployment.
A-Harness avoids task-specific model training and instead composes general-purpose skills through budgeted evidence acquisition and relative verification, making the same runtime transferable across domains.
The verifier adds explicit failure awareness---knowing \emph{when} evidence is insufficient and \emph{where} extra budget should be spent---which feed-forward predictors typically lack.
The two-tier memory further makes test-time scaling practical: a curated common-sense bank supplies broadly transferable priors, while an episodic bank accumulates verifier-accepted experience online, allowing the system to adapt as distribution shifts appear.
Together, these properties turn stronger foundation models into replaceable components of a self-correcting runtime rather than requiring a new affordance model for each domain.

\begin{table*}[t]
    \centering
    \caption{Quantitative results on ReasonAff and UMD datasets, comparing the performance of various foundation models as decision brains. Upper section: specialist and generalist models for reference; lower section: fixed-pipeline baselines and our adaptive-scheduling framework.}
    \vspace{-0.8em}
    \label{tab:comparison}
    \resizebox{\linewidth}{!}{
    \begin{tabular}{l|cccc|cccc}
        \toprule
        & \multicolumn{4}{c|}{\textbf{ReasonAff}} & \multicolumn{4}{c}{\textbf{UMD}} \\
        \textbf{Model}
            & \textbf{gIoU}$\uparrow$ & \textbf{cIoU}$\uparrow$ & \textbf{$P_{50}$}$\uparrow$ & \textbf{$P_{50-95}$}$\uparrow$
            & \textbf{gIoU}$\uparrow$ & \textbf{cIoU}$\uparrow$ & \textbf{$P_{50}$}$\uparrow$ & \textbf{$P_{50-95}$}$\uparrow$ \\
        \midrule
        VLPart~\cite{sun2023going}               & 4.21  & 3.88  & 1.31  & 0.85  & --    & --    & --    & --    \\
        OVSeg~\cite{liang2023open}               & 16.52 & 10.59 & 9.89  & 4.12  & --    & --    & --    & --    \\
        SAN~\cite{xu2023side}                    & 10.21 & 13.45 & 7.18  & 3.17  & --    & --    & --    & --    \\
        LISA-7B~\cite{lai2024lisa}               & 38.17 & 40.58 & 33.62 & 19.69 & 41.90 & 41.23 & 39.65 & 19.33 \\
        SAM4MLLM~\cite{chen2024sam4mllm}         & 45.51 & 33.64 & 43.48 & 22.79 & 12.40 & 8.41  & 4.12  & 0.05  \\
        AffordanceLLM~\cite{qian2024affordancellm} & 48.49 & 38.61 & 42.11 & 20.19 & 43.11 & 38.97 & 41.56 & 22.36 \\
        InternVL3-8B~\cite{zhu2025internvl3}     & 31.79 & 24.68 & 35.41 & 21.93 & --    & --    & --    & --    \\
        InternVL3-7B~\cite{zhu2025internvl3}     & --    & --    & --    & --    & 30.46 & 28.73 & 18.67 & 9.94  \\
        Qwen2.5VL-7B~\cite{qwen25vl}             & 25.18 & 20.54 & 26.00 & 15.82 & 33.21 & 29.83 & 25.17 & 10.45 \\
        AffordanceVLM~\cite{wu2025ragnet}         & 30.50 & 25.54 & 30.29 & 18.31 & 25.41 & 17.96 & 9.37  & 25.10 \\
        Seg-Zero~\cite{Liu2025SegZeroRG}          & 59.26 & 48.03 & 61.33 & 45.87 & 44.26 & 39.30 & 39.93 & 16.53 \\
        Vision Reasoner~\cite{Liu2025VisionReasonerUV} & 63.04 & 52.70 & 67.33 & 47.23 & 44.00 & 39.71 & 39.04 & 16.10 \\
        Affordance-R1~\cite{wang2025affordance}   & 67.41 & 62.72 & 74.50 & 55.22 & 49.85 & 42.24 & 53.35 & 34.08 \\
        ConverSeg~\cite{sahoo2026conversational} & 30.11 & 25.08 & 30.50 & 17.02 & 33.27 & 10.37 & 32.63 & 13.59 \\
        \midrule
        \multicolumn{9}{l}{\textit{w/o A-Harness}} \\[-1pt]
        \cmidrule(lr){1-9}
        Only w/ Det. \& Seg. skills                     & 51.86 & 43.73 & 57.00 & 38.30 & 46.53 & 37.77 & 53.24 & 30.56 \\
        Full Fixed Skill Chain                   & 55.05 & 49.57 & 58.07 & 37.48 & 50.19 & 49.24 & 55.88 & 29.75 \\
        \midrule
        \multicolumn{9}{l}{\textit{w/ A-Harness (Ours)}} \\[-1pt]
        \cmidrule(lr){1-9}
        w/ Qwen-3.5-397B-A17B~\cite{qwen3.5}              & 58.51 & 49.47 & 64.83 & 44.73 & \textbf{57.61} & 53.39 & \textbf{67.71} & \textbf{37.44} \\
        w/ Gemini-3-flash~\cite{deepmind_gemini_3_flash_modelcard} & 58.27 & 47.25 & 63.68 & 45.91 & 51.33 & 46.49 & 53.60 & 28.17 \\
        w/ GPT-4o~\cite{gpt4}                              & 60.53 & 54.91 & 66.73 & 45.53 & 52.74 & 50.04 & 57.62 & 29.85 \\
        w/ GLM-5~\cite{zeng2026glm}                        & 60.72 & 55.02 & 66.78 & 45.44 & 54.28 & 54.06 & 61.76 & 33.94 \\
        w/ Claude-Sonnet-4.6~\cite{anthropic_sonnet_46_systemcard} & 66.48 & 62.82 & 73.19 & 53.38 & 53.72 & 51.15 & 62.50 & 33.31 \\
        w/ Claude-Opus-4.6~\cite{anthropic_opus_46_systemcard} & \textbf{69.68} & \textbf{70.88} & \textbf{77.50} & \textbf{56.35} & 54.94 & \textbf{55.04} & 64.67 & 36.80 \\
        \bottomrule
    \end{tabular}
    }
\end{table*}

\section{Experiment}
\subsection{Experiment Settings}
\subsubsection{Datasets.}
We evaluate on three benchmarks covering diverse affordance reasoning scenarios: ReasonAff~\cite{wang2025affordance} (600 test image--task pairs, reasoning-oriented), UMD Part Affordance~\cite{UAD} (1,922 test images across 17 categories and 7 affordance types), and RAGNet~\cite{wu2025ragnet} (3,018 pairs across RAGNet-3DOI and RAGNet-HANDAL subsets). Dataset details are shown in Appendix~\ref{app:dataset_details}.
\vspace{-1em}
\subsubsection{Implementation Details.}
We construct the common-sense bank from the ReasonAff and UMD training splits, with selected RAGNet samples used for out-of-domain memory analysis, and evaluate only on the corresponding held-out test sets. GPT-4o~\cite{gpt4} serves as the default decision brain, with SAM2-Large~\cite{ravi2024sam} for segmentation and Qwen3-VL-235B-A22B~\cite{bai2025qwen3} for target-point reasoning. All results are averaged over three randomized test-set orderings to control for the order-dependent updates of $\mathcal{M}^{\textsc{tt}}$. Appendix~\ref{app:implementation} and Appendix~\ref{app:experiments} provide per-run variance, alternative tool/backbone configurations, hyperparameters, sensitivity analyses, and additional ablations.
\vspace{-1em}
\subsubsection{Baseline.}
We compare A-Harness against baselines spanning open-vocabulary segmentation, instruction-conditioned grounding, affordance-specialized models, and general-purpose reasoning LMMs; full descriptions and configurations are in Appendix~\ref{app:baselines}.

\vspace{-1em}
\subsection{Quantitative Analysis}
\vspace{-0.5em}
Table~\ref{tab:comparison} shows that A-Harness improves over fixed-pipeline baselines across both benchmarks, indicating that adaptive routing and verification-gated retries matter beyond the choice of backbone. Among decision brains, Claude-Opus-4.6 performs best on ReasonAff (69.68 gIoU and 70.88 cIoU), whereas Qwen-3.5 leads on UMD (57.61 gIoU and 67.71 $P_{50}$), suggesting that reasoning-heavy and category-diverse settings favor different model strengths. Even the weakest A-Harness variant surpasses specialist and generalist baselines on ReasonAff and remains competitive on UMD, supporting the claim that the main gain comes from orchestration rather than a single foundation model.

\vspace{-1em}
\subsection{Ablation study}
\vspace{-0.5em}
As shown in Table~\ref{tab:ablation}, stronger backbones (Qwen-3.5 for detection, SAM-3 for segmentation) yield clear gains, underscoring the dependence on base-model grounding quality. Auxiliary skills (web search, interaction-imagination) primarily boost localization precision, as reflected by the cIoU and $P_{50}$ decline when removed. Disabling the Verifier while keeping adaptive routing causes a consistent drop; notably, this variant even underperforms the two-skill (Det.\,\&\,Seg.) baseline on ReasonAff, showing that richer skill sets amplify the need for verification---without conflict resolution, additional evidence sources introduce noise that outweighs their benefit. Among memory components, $\mathcal{M}^{TT}$ has the largest impact via adaptive transfer to out-of-domain samples, while $\mathcal{M}^{CS}$ supplies complementary knowledge priors, validating our dual-memory design.

\begin{table}[!t]
    \centering
    \caption{Ablation study of A-Harness on ReasonAff and UMD datasets.
    \textit{Standard Setting} uses GPT-4o as the decision brain.
    \textit{Backbone Substitution} replaces the detection and segmentation backbone while keeping the full pipeline.
    \textit{Component Ablation} removes individual modules from the standard setting.}
    \vspace{-0.8em}
    \label{tab:ablation}
    \resizebox{\linewidth}{!}{
    \begin{tabular}{l|cccc|cccc}
        \toprule
        & \multicolumn{4}{c|}{\textbf{ReasonAff}} & \multicolumn{4}{c}{\textbf{UMD}} \\
        \textbf{Configuration}
            & \textbf{gIoU}$\uparrow$ & \textbf{cIoU}$\uparrow$ & \textbf{$P_{50}$}$\uparrow$ & \textbf{$P_{50-95}$}$\uparrow$
            & \textbf{gIoU}$\uparrow$ & \textbf{cIoU}$\uparrow$ & \textbf{$P_{50}$}$\uparrow$ & \textbf{$P_{50-95}$}$\uparrow$ \\
        \midrule
        Full Standard Config                             & 60.53 & 54.91 & 66.73 & 45.53 & 52.74 & 50.04 & 57.62 & 29.85 \\
        \cmidrule(lr){1-9}
        w/ Rex-omni~\cite{jiang2025detect}                 & 55.44 & 52.73 & 58.70 & 41.54 & 50.37 & 41.78 & 51.84 & 27.05 \\
        w/ Qwen2.5VL-7B-Instruct~\cite{qwen25vl}                           & 54.13 & 51.12 & 58.06 & 36.67 & 39.09 & 38.57 & 36.11 & 16.06 \\
        w/ Qwen-3.5-397B-A17B~\cite{qwen3.5}       & \textbf{62.56} & 56.27 & \textbf{68.11} & \textbf{47.98} & \textbf{54.15} & \textbf{53.82} & \textbf{63.18} & \textbf{33.56} \\
        w/ SAM-3~\cite{carion2025sam}                                           & 61.90 & \textbf{56.36} & 67.33 & 46.12 & 53.18 & 50.56 & 58.28 & 30.85 \\
        \cmidrule(lr){1-9}
        Only w/ Det. \& Seg. skill                               & 59.68 & 49.29 & 64.10 & 44.20 & 50.82 & 41.92 & 51.76 & 26.44 \\
        w/o Verifier (Router only) & 52.28 & 43.15 & 55.88 & 34.75 & 50.87 & 42.00 & 54.48 & 26.72 \\
        w/o $\mathcal{M}^{CS}$                             & 54.43 & 53.38 & 62.55 & 39.06 & 50.31 & 43.12 & 54.96 & 27.39 \\
        w/o $\mathcal{M}^{TT}$                             & 50.85 & 41.89 & 56.11 & 34.40 & 49.50 & 41.04 & 52.56 & 26.61 \\
        w/o $\mathcal{M}^{CS}$ \& $\mathcal{M}^{TT}$      & 49.41 & 39.05 & 43.29 & 22.18 & 48.98 & 42.10 & 49.36 & 25.24 \\
        \bottomrule
    \end{tabular}
    }
\end{table}
\begin{figure}[!t]
    \centering
    \includegraphics[width=1\linewidth]{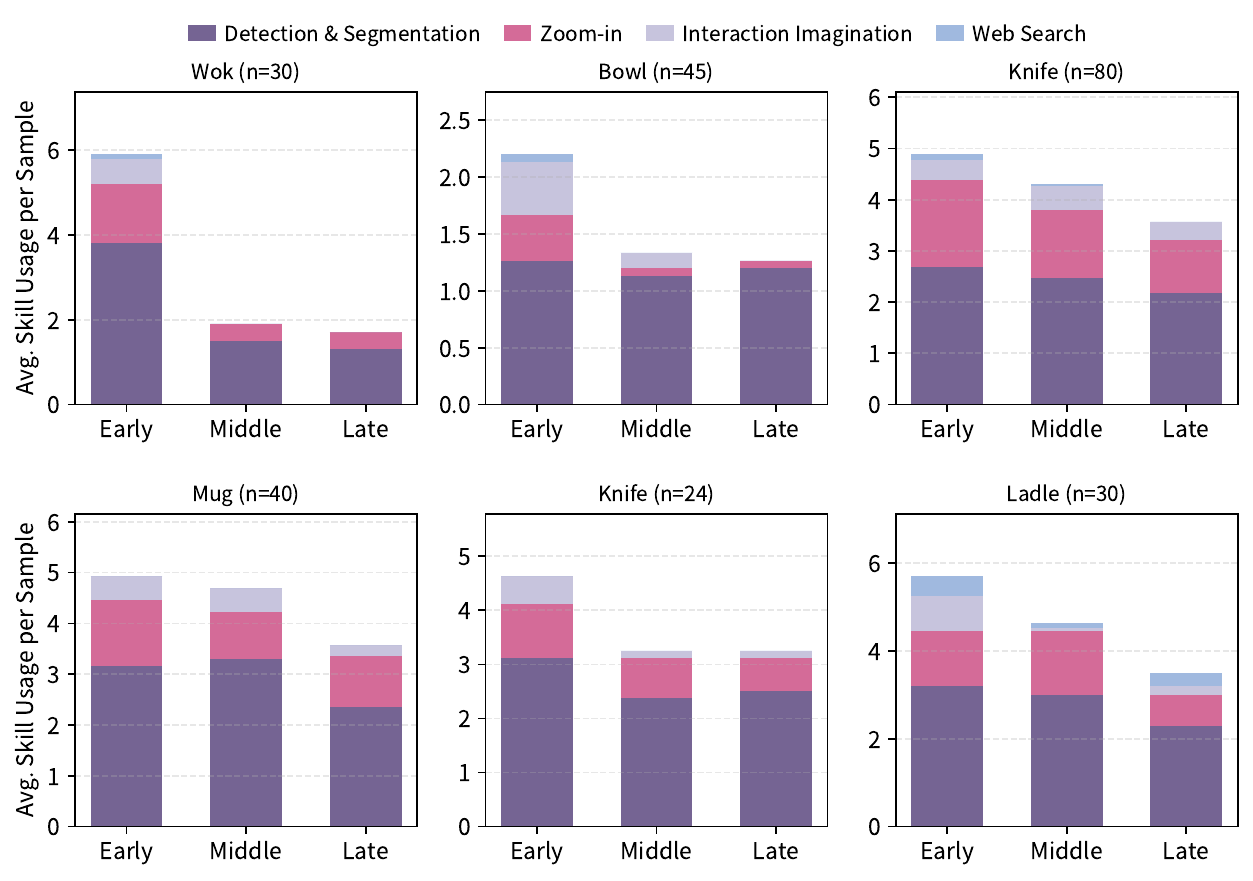}
    \vspace{-1.8em}
    \caption{Skill usage analysis on 3DOI (top) and UMD (bottom). Bars show average skill invocations per sample in early, middle, and late thirds of inference. A detailed efficiency comparison against fixed-pipeline baselines is in Appendix~\ref{app:efficiency}.}
    \label{fig:skill}
    \vspace{-2em}
\end{figure}

\vspace{-1em}
\subsection{Further analysis}
\vspace{-0.5em}
We next analyze how skill usage evolves, how the common-sense bank scales, and how the source/difficulty of $\mathcal{M}^\text{CS}$ affects cross-domain transfer. Unless otherwise stated, these experiments use GPT-4o~\cite{gpt4} as the decision brain, Qwen3-VL-235B-A22B~\cite{bai2025qwen3} for detection, and SAM2-LARGE~\cite{ravi2024sam} for segmentation.
\vspace{-0.8em}
\subsubsection{Analyses on skills usage.}
Fig.~\ref{fig:skill} shows that skill usage decreases from early to late inference stages as the episodic bank accumulates accepted trajectories. Detection and segmentation remain the dominant skills because they provide spatially grounded evidence, while interaction imagination and web search are invoked selectively when the verifier detects semantic ambiguity or insufficient support. This pattern is consistent with the intended role of the Router: auxiliary skills are reserved for cases where their expected diagnostic value justifies the added cost.
\vspace{-0.8em}
\subsubsection{Sensitivity Analysis on Common-Sense Bank.}
\begin{figure}[!t]
    \centering
    \includegraphics[width=\textwidth]{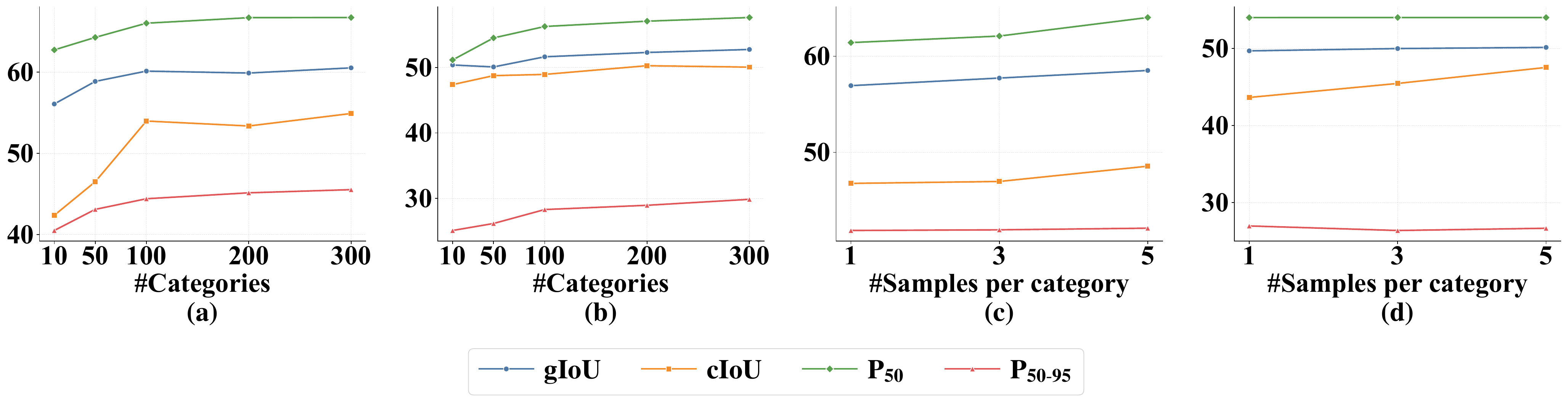}
    \vspace{-1.8em}
    \caption{Sensitivity analysis of the common-sense bank on ReasonAff and UMD datasets. (a)(b)~Performance under different numbers of categories. (c)(d)~Performance under different numbers of samples per category.}
    \label{fig:scaling_analysis}
\end{figure}

We vary both the number of categories and the number of samples per category in the common-sense bank. Increasing category coverage improves all metrics on ReasonAff and UMD (Fig.~\ref{fig:scaling_analysis}(a)(b)), but the gains flatten beyond roughly 200 categories, suggesting that additional categories become less useful once common functional patterns are covered. Denser sampling within each category (Fig.~\ref{fig:scaling_analysis}(c)(d)) primarily improves $cIoU$, indicating that repeated exemplars help the system stabilize larger affordance regions rather than only detecting small contact points.
\vspace{-0.8em}

\begin{table}[!t]
    \centering
    \caption{%
    \textbf{Zero-shot comparison with baselines and ablation on $\mathcal{M}^\text{CS}$.}
    \textit{Top block:} zero-shot baselines.
    \textit{Bottom blocks:} ablation over $\mathcal{M}^\text{CS}$ source (\textbf{a}--\textbf{c}) and exemplar difficulty (\textbf{d}--\textbf{g}).
    \textbf{a}/\textbf{b}/\textbf{c}: banks from UMD, RAGNet, and their 1\,:\,1 mixture;
    \textbf{d}/\textbf{e} and \textbf{f}/\textbf{g} pair HANDAL-easy/-hard exemplars with GPT-4o~\cite{gpt4} and Claude-Sonnet-4.6~\cite{anthropic_sonnet_46_systemcard}, respectively.
    A-Harness (w/o $\mathcal{M}^{\text{CS}}$) means the common-sense bank is empty.
    }
    \vspace{-0.8em}
    \label{tab:comprehensive}
    \adjustbox{max width=\textwidth}{
    \begin{tabular}{l|cc|cc|cc}
        \toprule
        \multirow{2}{*}{\textbf{Method / Setting}} & \multicolumn{2}{c|}{\textbf{3DOI}} & \multicolumn{2}{c|}{\textbf{HANDAL-easy}} & \multicolumn{2}{c}{\textbf{HANDAL-hard}} \\
        \cmidrule(lr){2-3} \cmidrule(lr){4-5} \cmidrule(lr){6-7}
         & \textbf{gIoU}$\uparrow$ & \textbf{cIoU}$\uparrow$ & \textbf{gIoU}$\uparrow$ & \textbf{cIoU}$\uparrow$ & \textbf{gIoU}$\uparrow$ & \textbf{cIoU}$\uparrow$ \\
        \midrule
        G-DINO~\cite{liu2024grounding} & 4.1  & 3.9  & 3.6  & 3.0  & 3.4  & 3.1  \\
        LISA~\cite{lai2024lisa}                                             & 12.3 & 8.1  & 15.5 & 11.9 & 12.3 & 8.1  \\
        GLaMM~\cite{rasheed2024glamm} & 4.4  & 2.9  & 4.7  & 3.5  & 5.0  & 3.5  \\
        Vision-Reasoner~\cite{Liu2025VisionReasonerUV} & 39.6 & 30.3 & 29.6 & 19.8 & 27.7 & 16.7 \\
        Affordance-R1~\cite{wang2025affordance} & 39.0 & 33.4 & 43.1 & 38.7 & 40.7 & 37.9 \\
        AffordanceVLM~\cite{wu2025ragnet} & 38.1 & 39.4 & 58.3 & 58.1 & 58.2 & 57.8 \\
        \midrule
        A-Harness (w/o $\mathcal{M}^{\text{CS}}$)
        & 56.5 & 47.2 & 58.4 & 57.6 & 55.3 & 55.1 \\
        \midrule
        w/ $\mathcal{M}^\text{CS}$-a         & 57.9 & 50.9 & 62.4 & 62.1 & 49.8 & 47.4 \\
        w/ $\mathcal{M}^\text{CS}$-b      & 63.8 & 52.0 & 61.4 & \textbf{63.3} & 60.6 & 61.0 \\
        w/ $\mathcal{M}^\text{CS}$-c       & 59.3 & 45.4 & 62.7 & 60.4 & \textbf{62.8} & \textbf{61.7} \\
        \midrule
        w/ $\mathcal{M}^\text{CS}$-d & 64.1 & 52.8 & 61.8 & 61.7 & 48.0 & 42.5 \\
        w/ $\mathcal{M}^\text{CS}$-e & 61.2 & 51.3 & 59.1 & 61.9 & 57.9 & 56.5 \\
        w/ $\mathcal{M}^\text{CS}$-f & \textbf{65.6} & \textbf{53.7} & \textbf{63.5} & 61.8 & 55.2 & 49.7 \\
        w/ $\mathcal{M}^\text{CS}$-g & 62.2 & 52.1 & 60.2 & 58.9 & 59.4 & 59.1 \\

        \bottomrule
    \end{tabular}
    }
    \vspace{-0.8em}
\end{table}

\vspace{-0.4em}
\subsubsection{Ablation on $\mathcal{M}^\text{CS}$: Common-Sense Source and Exemplar Difficulty.}
\label{sec:ablation_cs}

To isolate the role of common-sense memory, we evaluate on the RAGNet-family benchmarks (3DOI and HANDAL) in a zero-shot setting while varying the source and difficulty of $\mathcal{M}^\text{CS}$.

Table~\ref{tab:comprehensive} shows that the empty-bank variant is already competitive, while memory quality changes transfer behavior. UMD exemplars (\textbf{a}) help HANDAL-easy but not HANDAL-hard; RAGNet exemplars (\textbf{b}) transfer more uniformly; and the mixture (\textbf{c}) improves HANDAL-hard yet lowers 3DOI cIoU, suggesting that source mismatch can inject retrieval noise. Difficulty has a complementary effect: easy exemplars (\textbf{d}/\textbf{f}) help out-of-domain transfer, whereas hard exemplars (\textbf{e}/\textbf{g}) support difficult in-domain cases. Thus memory construction should balance source coverage and exemplar hardness.
\vspace{-0.5em}
\paragraph{Case study and computational complexity.}
Appendix~\ref{app:case_study} visualizes representative success and failure cases, focusing on how the Verifier redirects the Router when intermediate evidence conflicts or remains insufficient.
Appendix~\ref{app:compute} reports the cost profile: most samples are resolved after one detection call, while additional calls are concentrated on hard cases; performance gains diminish beyond three calls, motivating the early-exit threshold used in our default budget.

\vspace{-0.5em}
\section{Conclusion}
\enlargethispage{3\baselineskip}
\vspace{-0.1em}
\label{sec:conclusion}

We presented Affordance Agent Harness, a closed-loop runtime that replaces fixed affordance pipelines with budgeted, per-instance skill orchestration.
A-Harness gates commitments with relative checks for cross-skill agreement, cross-scale stability, and evidence sufficiency; when a check fails, it triggers targeted recovery instead of passing errors to late fusion.
Its two-tier memory amortizes successful experience across recurring objects and episodes, turning prior trajectories into reusable action priors.
Across benchmarks, this improves the accuracy--cost trade-off by adding skills only when evidence is insufficient, showing that open-world visual reasoning needs runtimes that decide what to acquire, when to stop, and how to recover from unreliable outputs, not only stronger individual models.

\clearpage
\appendix
\renewcommand{\theHsection}{appendix.\Alph{section}}
\input{appendix}


%
%
\bibliographystyle{splncs04}
\bibliography{main}

\end{document}

%% file: appendix.tex
\section{Additional Methodology}
\label{app:methodology}
\subsection{Details on choosing $w_{\tau}$}
\label{app:w_tau}

In practice, $w_{\tau}$ follows a coarse reliability prior: evidence that is \emph{spatially grounded} and \emph{directly produced in the image frame} (e.g., high-resolution segmentation under a verified ROI) receives higher weight, while evidence that is \emph{indirect} or \emph{noisy} (e.g., web-retrieved text/images with potential domain mismatch) receives lower weight.
Accordingly, we use a coarse initialization such as
\begin{equation}
w_{\textsc{seg}} > w_{\textsc{det}} > w_{\textsc{dream}} \gtrsim w_{\textsc{web}},
\label{eq:w_prior}
\end{equation}
which reflects the empirical property that some tools are systematically noisier: for instance, web search often retrieves irrelevant or style-mismatched content, thus it should contribute to sufficiency only when corroborated by grounded visual evidence.

In all reported experiments, we initialize $w_{\textsc{seg}}=0.35$, $w_{\textsc{det}}=0.30$, $w_{\textsc{dream}}=0.20$, and $w_{\textsc{web}}=0.15$. At each step $t$, these weights are modulated by three multiplicative reliability factors:
\begin{equation}
\widetilde{w}_{\tau}^{(t)} = w_{\tau} \cdot \rho_{\tau}^{(t)} \cdot \phi_{\tau}^{(t)} \cdot \psi_{\tau}^{(t)},
\label{eq:w_dynamic}
\end{equation}
where $\rho_{\tau}^{(t)} \in [0,1]$ is the tool's self-confidence (e.g., SAM2's mask score), $\phi_{\tau}^{(t)}$ indicates cross-tool corroboration, and $\psi_{\tau}^{(t)} = \mathbb{1}[\zeta_{\tau}^{(t)} \ge 0.7]$ indicates cross-scale stability. For spatial evidence, we instantiate corroboration as $\phi_{\tau}^{(t)} = \mathbb{1}[\exists\, \xi' \in E_t \setminus \{\xi\}: \mathrm{IoU}(\xi, \xi') \ge 0.5]$; for non-spatial evidence such as retrieved text or web cues, we instead use semantic agreement with the currently verified visual evidence and instruction. The updated weights are re-normalized to sum to one before use in the sufficiency computation. This ensures that well-corroborated, stable evidence retains its full weight, while uncorroborated or drifting outputs (e.g., noisy web retrieval) are suppressed toward zero. Beyond this hand-crafted modulation, one could further let the Verifier/Router predict $w_{\tau}$ from richer tool metadata and reliability signals, but this optional extension is not used in the reported results.

\subsection{Full Instantiation of $\widehat{\Delta v}$}
\label{app:delta_v}

We instantiate $\widehat{\Delta v}(u; z_t)$ via a two-stage procedure. 

\textbf{Verifier-Gap Heuristic.} When commitment is denied, the Verifier identifies the dominant deficiency $\ell_t \in \{\omega, \zeta, \mu\}$ and proposes a corrective action $u_{\mathrm{prop}} = \Gamma(\ell_t)$: $\omega$-failure triggers ROI refinement and re-segmentation; $\zeta$-failure triggers $2\times$ zoom-in with re-detection; $\mu$-failure invokes a complementary unused skill (Dreamer for familiar categories, web search otherwise). For each candidate $u$, the estimated gain is:
\begin{equation}
\widehat{\Delta v}(u; z_t) =
\begin{cases}
    \lambda_{\ell_t} \cdot (\tau_{\ell_t} - \ell_t^{(t)})_+ & \text{if } u = u_{\mathrm{prop}}, \\[4pt]
    \lambda_{\ell'} \cdot (\tau_{\ell'} - \ell'^{(t)})_+ \cdot \eta_{\text{off}} & \text{otherwise},
\end{cases}
\label{eq:delta_v}
\end{equation}
where $\lambda_{\ell}$ are dimension-specific scales ($\lambda_\omega{=}1.0$, $\lambda_\zeta{=}0.8$, $\lambda_\mu{=}0.6$), $\tau_{\ell}$ denotes the retry threshold for diagnostic dimension $\ell$ (with $\tau_{\omega}=\underline{\omega}$ and analogous thresholds for $\zeta$ and $\mu$), and $\eta_{\text{off}}{=}0.5$ discounts off-target actions, so the Verifier's proposal receives the highest expected gain while the Router may override it for a more cost-effective alternative.

\textbf{LLM Fallback.} When the top-2 candidates are near-tied ($<$10\% gap) or the same deficiency persists for two consecutive retries, the Router delegates to the LLM decision brain, which receives the evidence summary $\psi(E_t)$, diagnostic scores, available skills, and retrieved memory entries, and outputs a JSON-formatted action with a rationale. In practice, ${\sim}$85\% of routing decisions are resolved by the heuristic alone.

\section{Additional Experiments}
\label{app:experiments}
\subsection{Experiment Settings}
\subsubsection{Dataset Details.}
\label{app:dataset_details}
We evaluate A-Harness on three primary benchmarks, each targeting different aspects of affordance reasoning and grounding:

\begin{itemize}
    \item \textbf{ReasonAff~\cite{wang2025affordance}:} This dataset focuses on reasoning-heavy affordance tasks. It contains complex instructions that require the model to understand the functional intent behind an interaction (\eg, ``find a place to hold the cup safely''). We use the test split consisting of 600 high-quality image--task pairs.
    \item \textbf{UMD Part Affordance~\cite{UAD}:} A classic benchmark for part-level affordance. It covers 17 object categories (\eg, bowl, hammer, saw) and 7 affordance types (\eg, grasp, cut, scoop). We follow the standard sampling protocol, using 2,004 test images to ensure a robust evaluation of category-level generalization.
    \item \textbf{RAGNet~\cite{wu2025ragnet}:} A large-scale benchmark for reasoning-driven affordance. We evaluate on two subsets: \textbf{RAGNet-3DOI}, which focuses on 3D object interactions in diverse scenes, and \textbf{RAGNet-HANDAL}, which emphasizes hand-object alignment and tool usage. The combined test set includes 3,018 image--task pairs.
\end{itemize}

\subsubsection{Implementation Details.}
\label{app:implementation}
\paragraph{Hardware and Software Environment.}
All experiments are conducted on a server equipped with 8 $\times$ NVIDIA H100 (80GB) GPUs and an Intel(R) Xeon(R) Platinum 8480+ CPU. The software environment is based on Python 3.10, PyTorch 2.4.0, and CUDA 12.2. For API-based models used in our study, we use the official Python SDKs with a consistent temperature setting of 0.0 to ensure reproducibility.

\paragraph{Skill Configurations.}
\begin{itemize}
    \item \textbf{Detection:} We use Qwen3-VL-235B-A22B as the default detector. For each detection call, we provide the original image and the instruction $q$, requesting the model to output bounding boxes in $[x_{min}, y_{min}, x_{max}, y_{max}]$ format normalized to $[0, 1000]$.
    \item \textbf{Segmentation:} SAM2-Large is employed for mask generation. When the Router provides a point prompt, we use the highest-confidence mask output by SAM2. For box prompts, we use the bounding box to constrain the initial search area.
    \item \textbf{Web Search:} We utilize the Google Search API to retrieve the top-3 relevant text snippets and, if available, 2 reference images for the given object-affordance pair.
\end{itemize}
In the commit rule, the squashing function $\sigma(\cdot)$ is implemented as a sigmoid, mapping the model-produced sufficiency value to $[0,1]$ before aggregation into the commit score.

\paragraph{Memory.}
The common-sense bank $\mathcal{M}^{\textsc{cs}}$ is pre-populated with 200 categories from the training sets of ReasonAff and UMD. Each category contains 5 diverse exemplars. The retrieval embeddings are computed using a concatenated feature vector from DINO-v3-Large and SigLIP-SO400M, providing both geometric and semantic alignment.

\subsubsection{Baseline Configurations}
\label{app:baselines}
To ensure a rigorous evaluation, we compare A-Harness with representative baselines covering complementary capability sources:
\begin{itemize}
    \item \textbf{Open-Vocabulary Segmenters:} We include VLPart~\cite{sun2023going}, OVSeg~\cite{liang2023open}, and SAN~\cite{xu2023side}. These models rely on large-scale pre-training to segment objects from natural language descriptions but lack explicit affordance reasoning.
    \item \textbf{Instruction-Conditioned LMMs:} LISA~\cite{lai2024lisa} and GLaMM~\cite{rasheed2024glamm} are evaluated as representatives of models that output masks directly from text instructions. We use their respective 7B/13B variants with default prompts.
    \item \textbf{Affordance-Specialized Models:} We compare against AffordanceLLM~\cite{qian2024affordancellm}, AffordanceVLM~\cite{wu2025ragnet}, and the recent Affordance-R1~\cite{wang2025affordance}. These models are either fine-tuned on affordance datasets or use specialized reasoning chains.
    \item \textbf{Reasoning-Augmented Models:} Seg-Zero~\cite{Liu2025SegZeroRG}, VisionReasoner~\cite{Liu2025VisionReasonerUV}, and ConverSeg~\cite{sahoo2026conversational} are included to contextualize our performance against systems that incorporate richer semantic or multi-step reasoning for segmentation. In particular, ConverSeg targets conversational and abstract concept grounding, including affordances and functions, but remains a feed-forward segmentation model rather than a closed-loop agent.
\end{itemize}


\subsubsection{Evaluation Metrics.}
We evaluate affordance grounding with four complementary metrics following established protocols in affordance prediction~\cite{wang2025affordance,wan2025instructpart} and semantic segmentation~\cite{carion2025sam,zhong2025omnisam}:
\begin{itemize}
    \item \textbf{gIoU (Generalized IoU):} The mean Intersection-over-Union calculated across all test samples, reflecting the overall segmentation fidelity at the instance level.
    \item \textbf{cIoU (Cumulative IoU):} The ratio of the total intersection area to the total union area across the entire dataset, offering a global measure of prediction robustness.
    \item \textbf{P@50 (Precision at 0.5):} The proportion of samples where the prediction IoU exceeds the 0.5 threshold, evaluating the system's ability to achieve reliable localization.
    \item \textbf{P@50:95:} The average precision computed across a range of IoU thresholds from 0.50 to 0.95 (with a step of 0.05), providing a stringent evaluation of spatial alignment and boundary precision.
\end{itemize}

\subsubsection{Hyperparameter Configuration.}
Table~\ref{tab:hyperparams} summarizes the default hyperparameter configurations used in our experiments. These values are selected based on the sensitivity analyses presented in the following sections to achieve an optimal balance between grounding accuracy and computational cost.

\begin{table}[!t]
    \centering
    \caption{Default hyperparameter configurations for A-Harness.}
    \label{tab:hyperparams}
    \adjustbox{max width=\textwidth}{
    \begin{tabular}{c|c}
        \toprule
        \textbf{Hyperparameter} & \textbf{Value} \\
        \midrule
        $C_{\textsc{cs}}$ & 1000 \\
        $C_{\textsc{tt}}$ & 80 \\
        $N$ & 2 \\
        $B$ & 3 \\
        $\delta$ & 0.8 \\
        $\underline{\omega}$ & 0.5 \\
        $\alpha$ & 0.5 \\
        $\beta$ & 0.3 \\
        $\gamma$ & 0.2 \\
        \bottomrule
    \end{tabular}
    }
\end{table}

\subsubsection{Prompt of A-Harness.}
\begin{figure}[!tp]
    \centering
    \includegraphics[width=\textwidth]{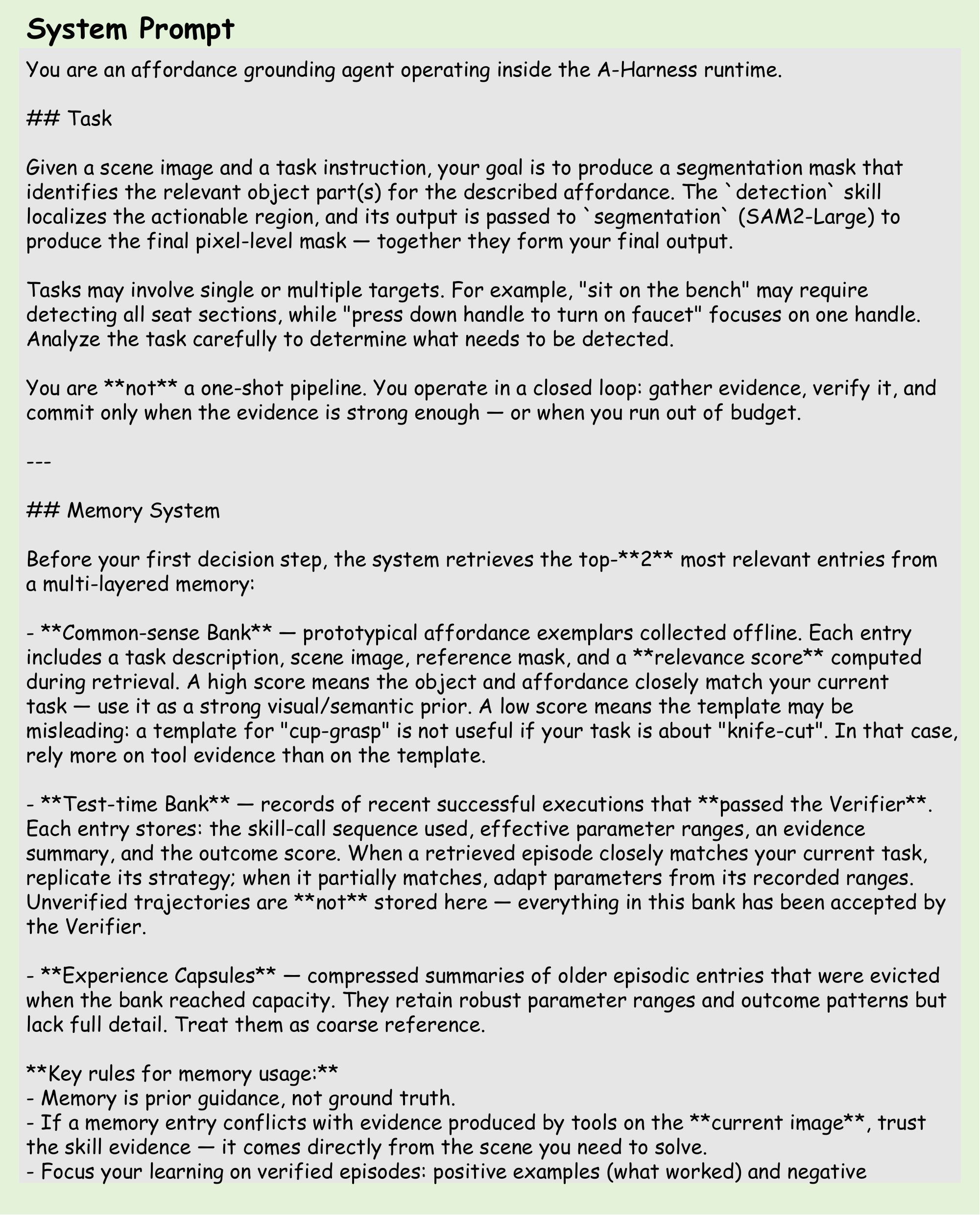}
    \vspace{-1.5em}
    \caption{The first page of the system prompt.}
    \label{fig:system_prompt_1}
\end{figure}

\begin{figure}[!tp]
    \centering
    \includegraphics[width=\textwidth]{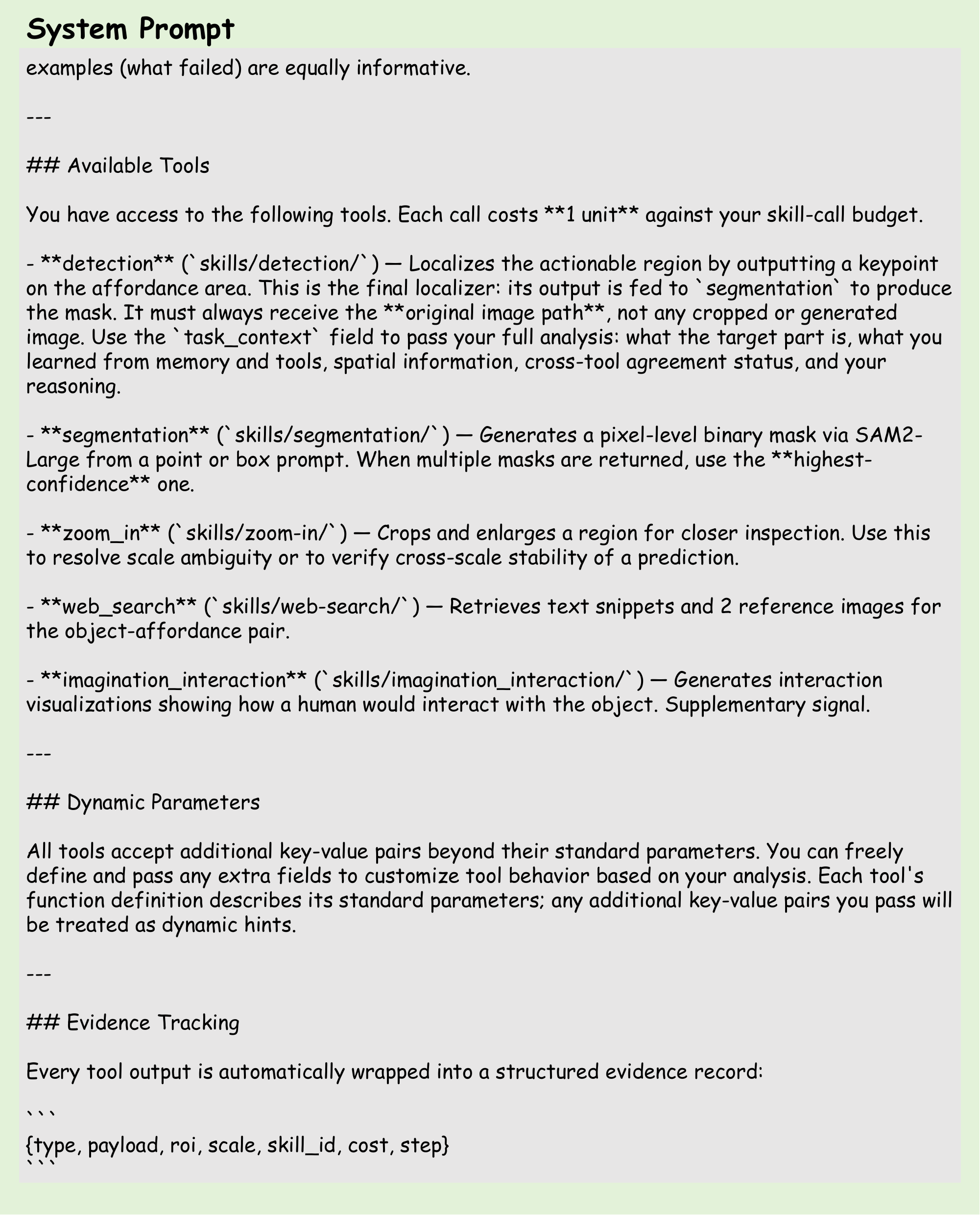}
    \vspace{-1.5em}
    \caption{The second page of the system prompt.}
    \label{fig:system_prompt_2}
\end{figure}

\begin{figure}[!tp]
    \centering
    \includegraphics[width=\textwidth]{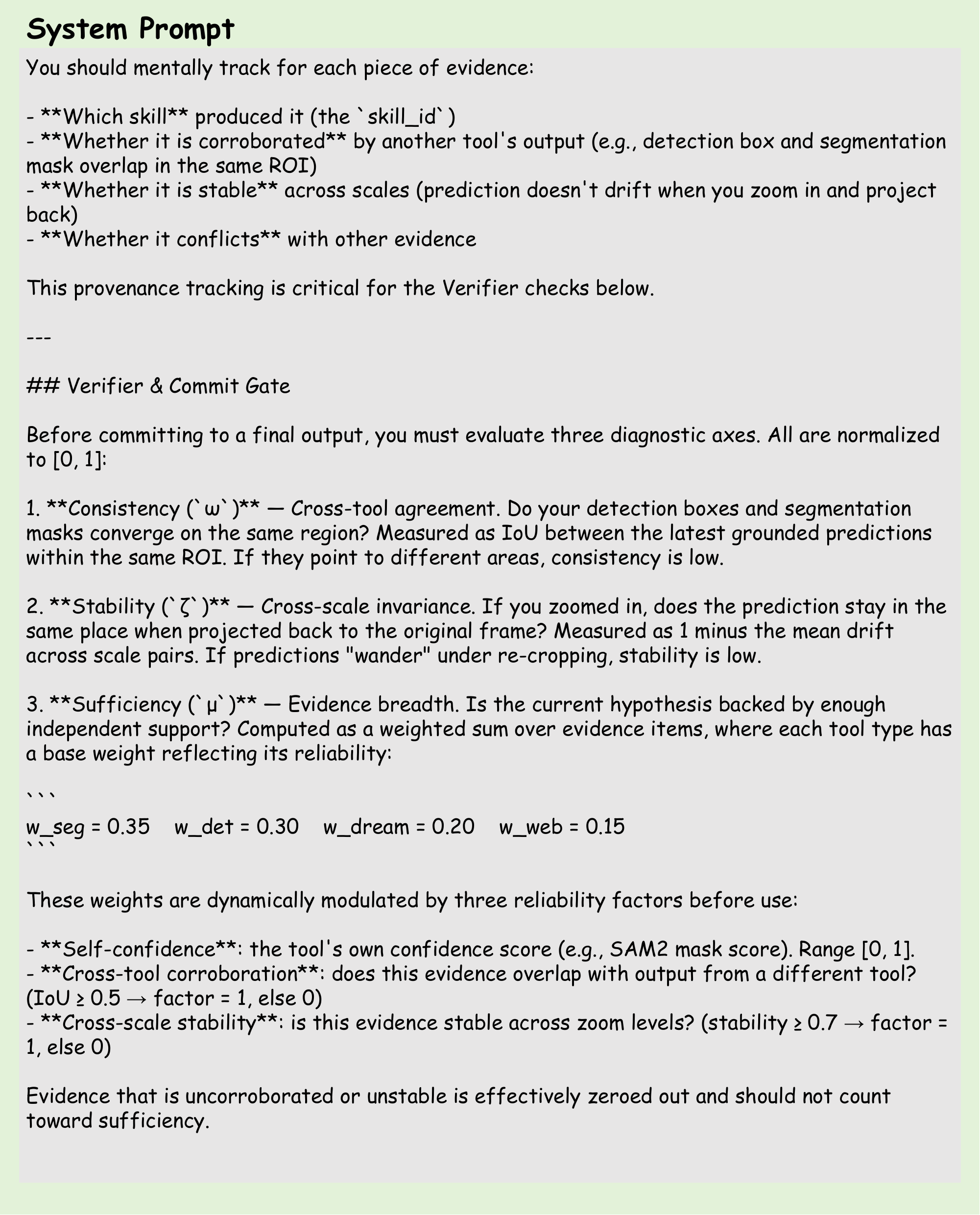}
    \vspace{-1.5em}
    \caption{The third page of the system prompt.}
    \label{fig:system_prompt_3}
\end{figure}

\begin{figure}[!tp]
    \centering
    \includegraphics[width=\textwidth]{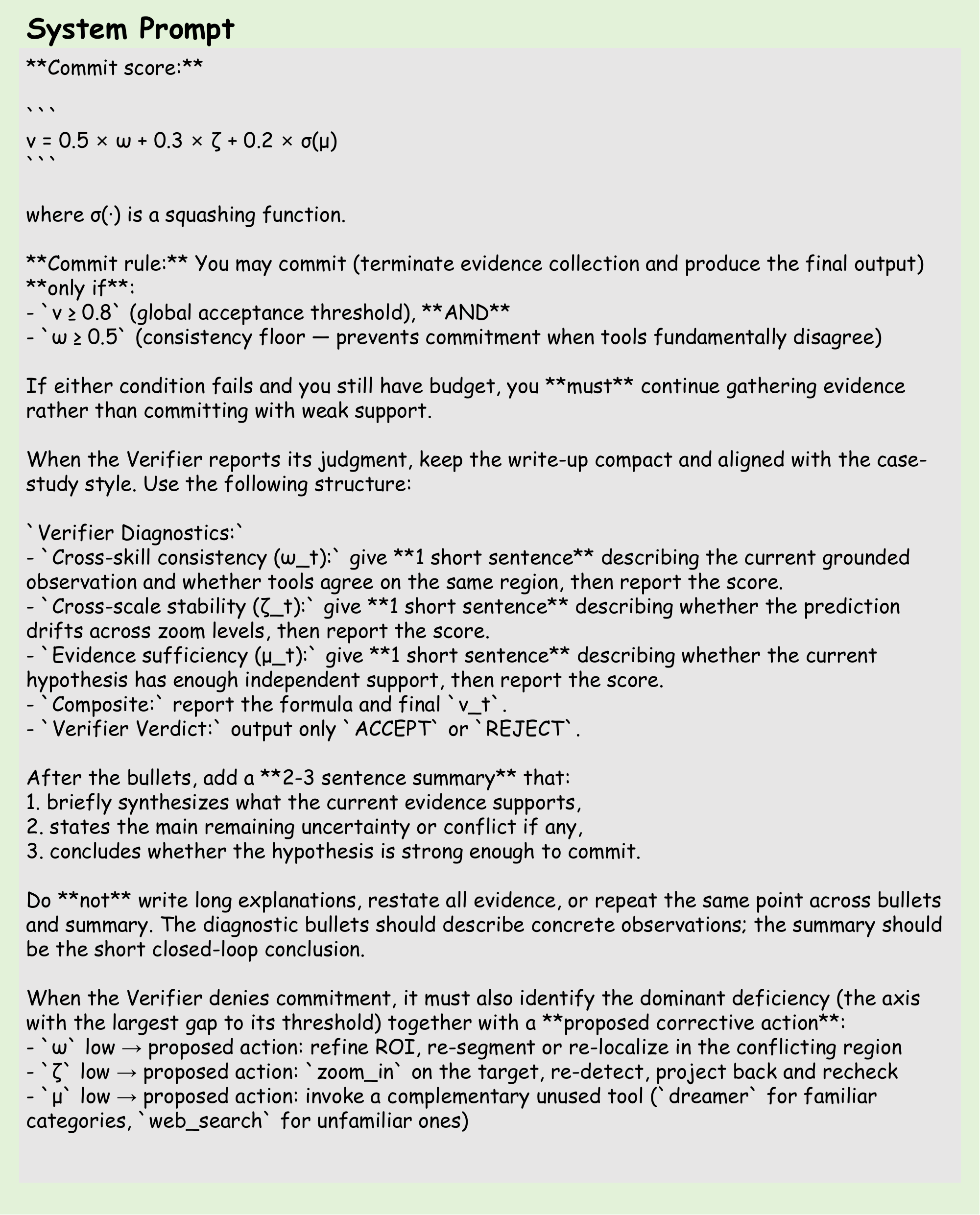}
    \vspace{-1.5em}
    \caption{The fourth page of the system prompt.}
    \label{fig:system_prompt_4}
\end{figure}

\begin{figure}[!tp]
    \centering
    \includegraphics[width=\textwidth]{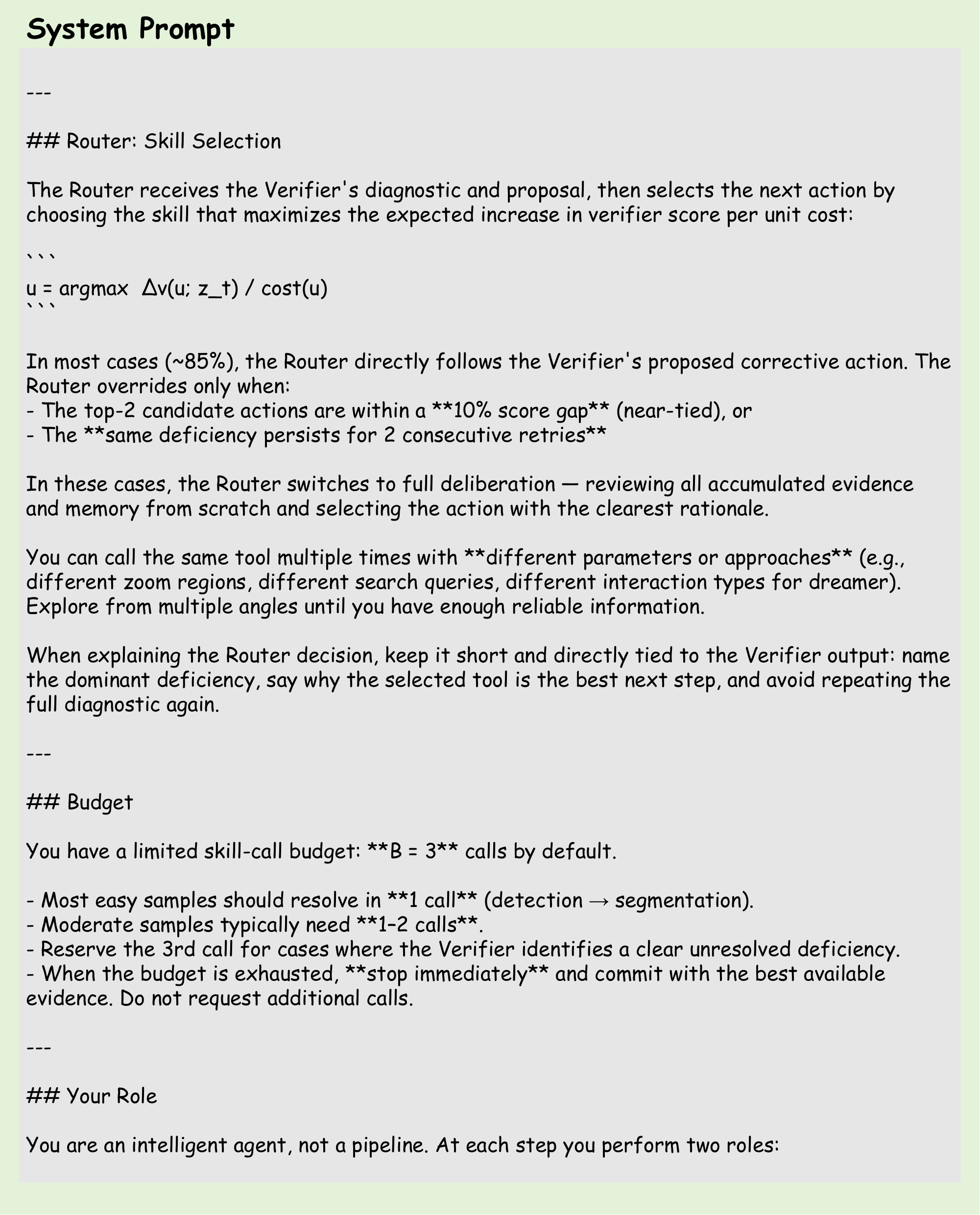}
    \vspace{-1.5em}
    \caption{The fifth page of the system prompt.}
    \label{fig:system_prompt_5}
\end{figure}

\begin{figure}[!tp]
    \centering
    \includegraphics[width=\textwidth]{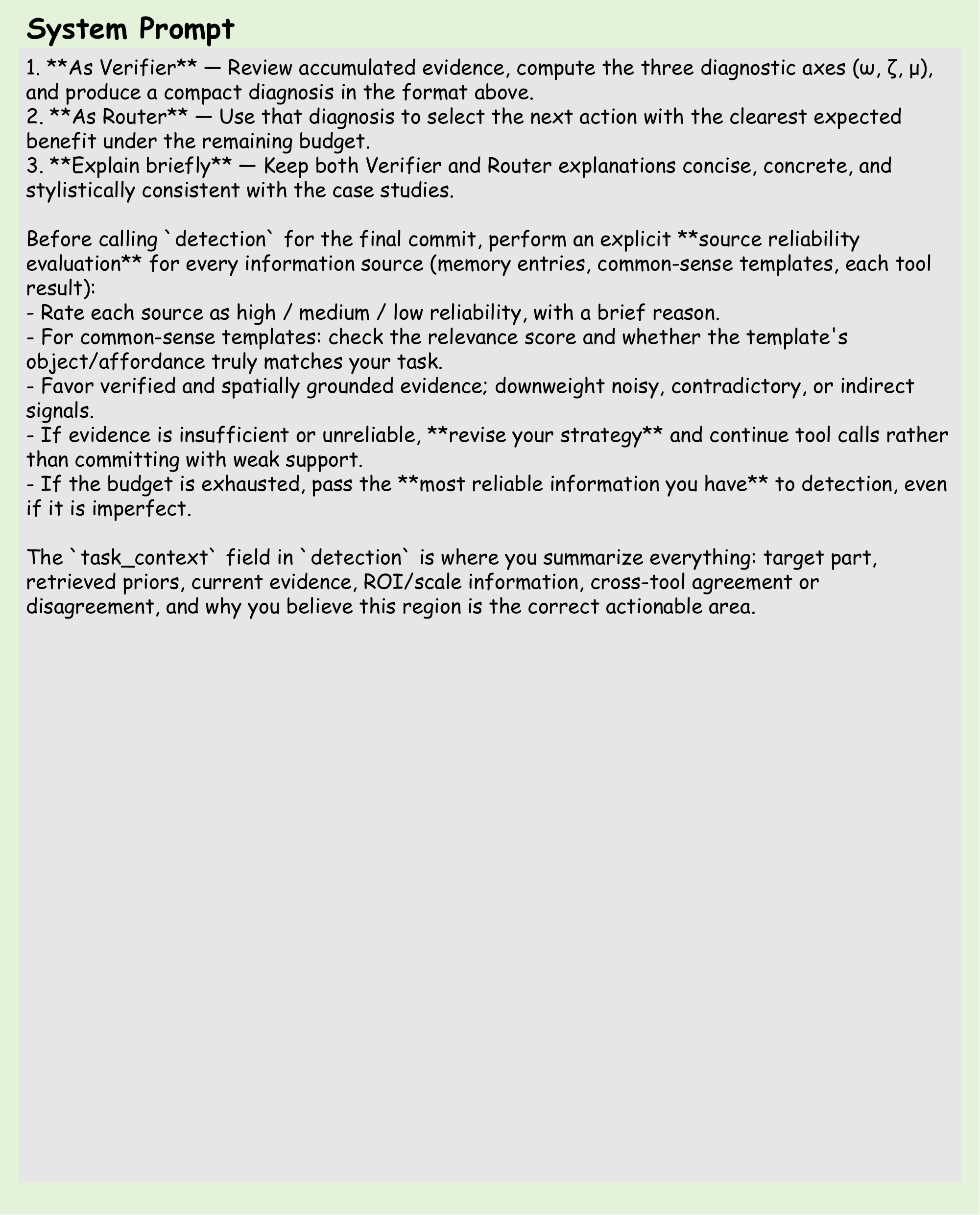}
    \vspace{-1.5em}
    \caption{The sixth page of the system prompt.}
    \label{fig:system_prompt_6}
\end{figure}

\begin{figure}[!tp]
    \centering
    \includegraphics[width=\textwidth]{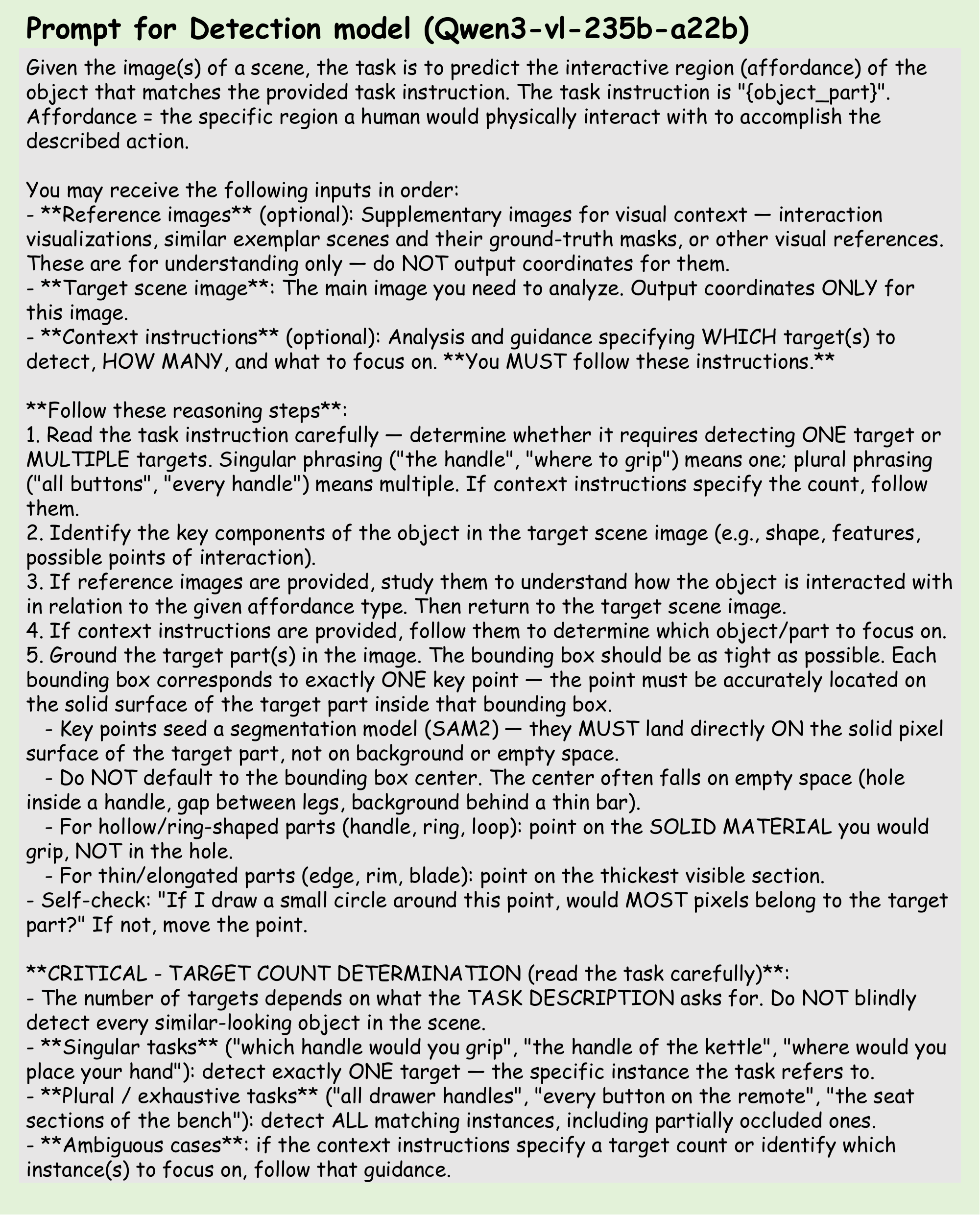}
    \vspace{-1.5em}
    \caption{The first page of the prompt for detection model.}
    \label{fig:prompt_for_detection_model_1}
\end{figure}

\begin{figure}[!tp]
    \centering
    \includegraphics[width=\textwidth]{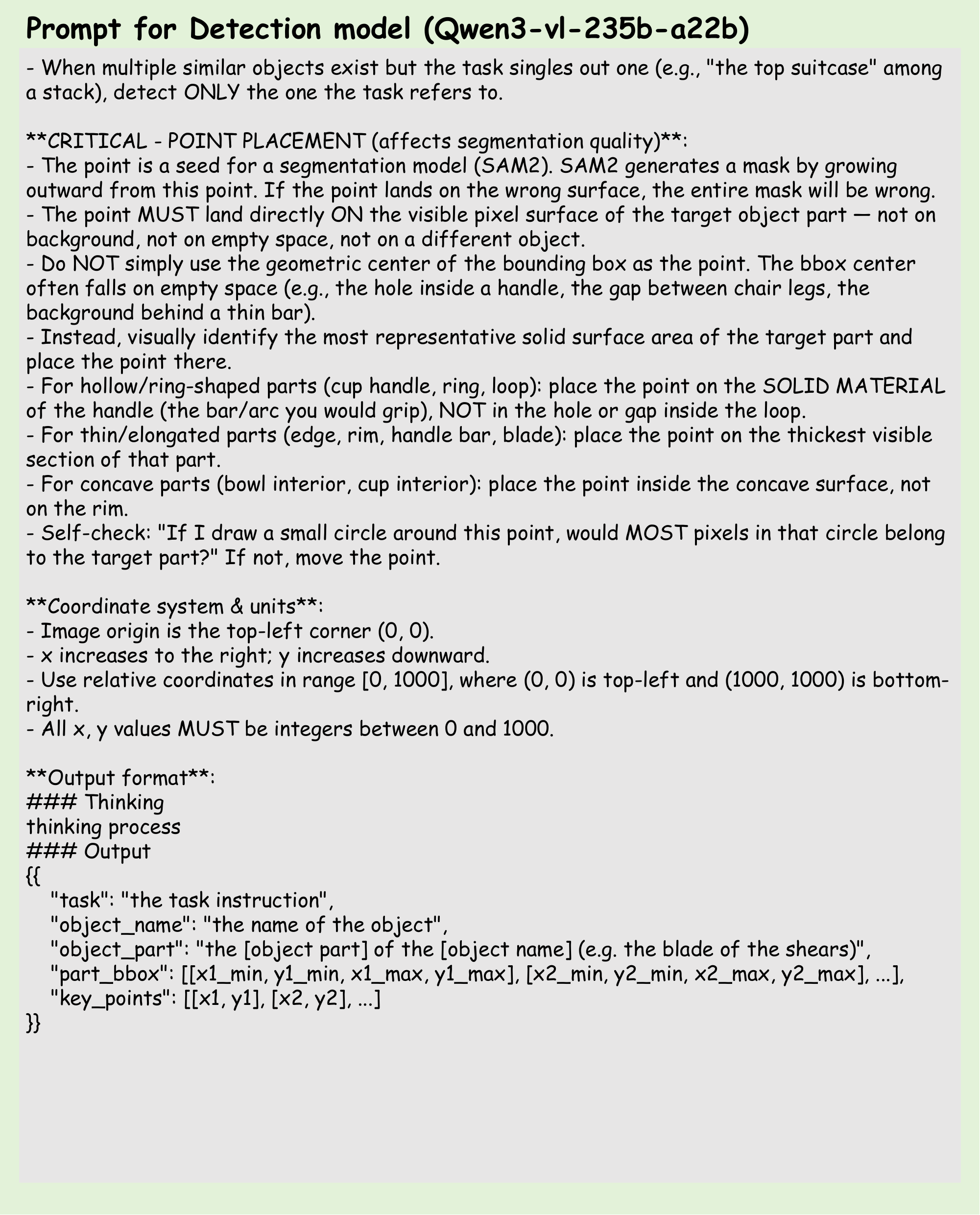}
    \vspace{-1.5em}
    \caption{The second page of the prompt for detection model.}
    \label{fig:prompt_for_detection_model_2}
\end{figure}

\begin{figure}[!tp]
    \centering
    \includegraphics[width=\textwidth]{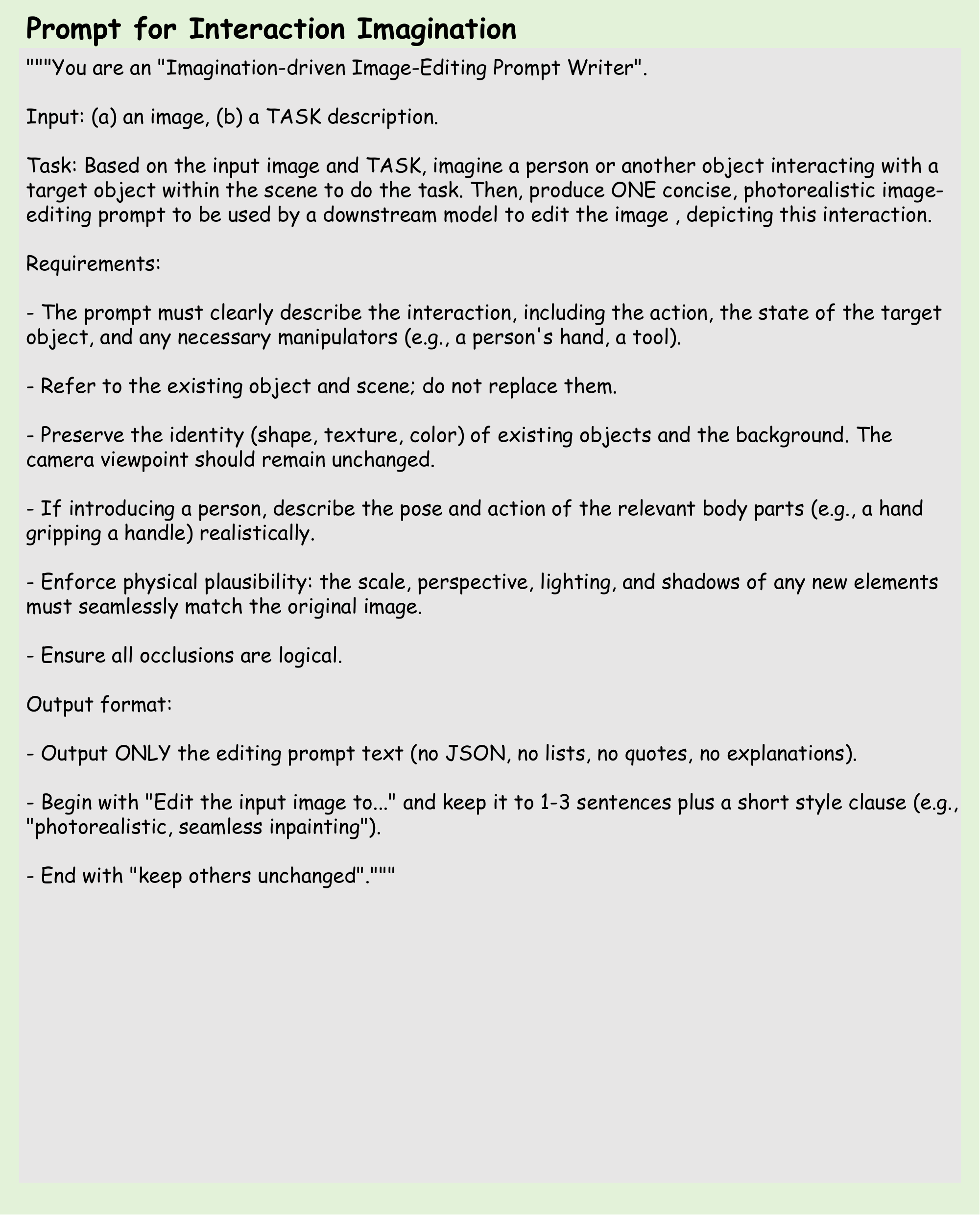}
    \vspace{-1.5em}
    \caption{The prompt for interaction imagination.}
    \label{fig:prompt_for_interaction_imagination}
\end{figure}

We provide the prompts used in A-Harness for completeness. The full system prompt is shown in Figs.~\ref{fig:system_prompt_1}--\ref{fig:system_prompt_6}. The prompt for the detection model is shown in Figs.~\ref{fig:prompt_for_detection_model_1} and \ref{fig:prompt_for_detection_model_2}, and the prompt for interaction imagination is shown in Fig.~\ref{fig:prompt_for_interaction_imagination}.

\subsection{Further analysis}
\subsubsection{Computational complexity analysis.}
\label{app:compute}
The majority of samples (\emph{e.g.}, 500/600 on ReasonAff, 1{,}127/2{,}004 on UMD) resolve within a single detection call, achieving strong grounding performance at minimal token cost, confirming that easy instances are handled efficiently without unnecessary computation. For harder samples requiring 2-3 retries, token consumption grows proportionally, demonstrating that the system adaptively allocates greater computational effort to more challenging instances. Unless otherwise stated, the main experiments use a skill-call budget of $B{=}3$; by contrast, the statistics in Table~\ref{tab:det_metrics_comparison} are collected \emph{without} truncating runs at this budget, so as to expose the full tail behavior of repeated detection retries.
\begin{table}[!t]
\centering
\caption{Computational cost and performance across detection retry counts on ReasonAff and UMD datasets. Each row reports the number of samples resolved at that retry count, the mean token consumption, and the corresponding gIoU/cIoU scores, with a horizontal rule separating samples solved within 3 retries from those requiring more. These statistics are computed without enforcing the default budget truncation $B{=}3$, and are included to show the unconstrained retry tail.}
\label{tab:det_metrics_comparison}
\adjustbox{max width=\textwidth}{%
\begin{tabular}{c|cc|cc|cc|cc}
\toprule
\multirow{2}{*}{\textbf{Det Calls}} & \multicolumn{2}{c|}{\textbf{N}} & \multicolumn{2}{c|}{\textbf{Mean Tokens}} & \multicolumn{2}{c|}{\textbf{gIoU}} & \multicolumn{2}{c}{\textbf{cIoU}} \\
\cmidrule(lr){2-3} \cmidrule(lr){4-5} \cmidrule(lr){6-7} \cmidrule(lr){8-9}
 & \textbf{ReasonAff} & \textbf{UMD} & \textbf{ReasonAff} & \textbf{UMD} & \textbf{ReasonAff} & \textbf{UMD} & \textbf{ReasonAff} & \textbf{UMD} \\
\midrule
1 & 500 & 1,127 & 4,716 & 4,648 & 0.7437 & 0.6720 & 0.7429 & 0.6768 \\
2 & 65 & 365 & 7,309 & 7,207 & 0.5193 & 0.4865 & 0.4472 & 0.4885 \\
3 & 21 & 219 & 11,081 & 9,855 & 0.3423 & 0.4213 & 0.3065 & 0.4588 \\
\hline
4 & 6 & 105 & 15,464 & 12,585 & 0.3067 & 0.3101 & 0.2964 & 0.3022 \\
5 & 5 & 77 & 17,279 & 15,622 & 0.3182 & 0.2680 & 0.2590 & 0.2608 \\
6 & 1 & 41 & 21,012 & 18,415 & 0.2625 & 0.1854 & 0.2625 & 0.1512 \\
7 & 2 & 58 & 23,439 & 20,652 & 0.2717 & 0.1502 & 0.2955 & 0.0935 \\
8 & - & 12 & - & 22,587 & - & 0.1245 & - & 0.0964 \\
\bottomrule
\end{tabular}%
}
\end{table}

\subsubsection{Efficiency comparison against fixed-pipeline baselines.}
\label{app:efficiency}
Table~\ref{tab:efficiency} compares our method against three fixed-workflow baselines on ReasonAff and UMD, reporting both accuracy (gIoU, cIoU) and efficiency metrics (average tokens, latency, and estimated API cost).
Among them, ``Det.\,\&\,Seg.'' and ``Full Fixed Chain'' are our own baselines: the former uses a minimal workflow with only detection and segmentation, while the latter executes all skills described in our framework in a fixed order for every sample.
For A4-Agent~\cite{zhang2025a4agent}, we include it as a closely related agentic affordance baseline.
A4-Agent represents a predefined skill-composition workflow centered on interaction imagination, detection, and segmentation, and therefore provides a natural reference point for evaluating agent-style affordance grounding.
Together with our Det.\,\&\,Seg. and Full Fixed Chain baselines, this comparison helps characterize how verification-gated adaptive routing affects the accuracy--efficiency trade-off.

As shown in Table~\ref{tab:efficiency}, our method achieves the best overall accuracy on both benchmarks while remaining substantially more efficient than heavier fixed workflows.
Compared with the Full Fixed Chain, ours reduces token usage, latency, and cost by a large margin, while also improving both gIoU and cIoU, showing that adaptive routing avoids unnecessary skill calls without sacrificing performance.
Compared with the lighter baseline that uses only detection and segmentation, our method introduces only moderate overhead but yields clear gains in grounding quality, indicating that verification and selective retries mainly spend extra computation on genuinely difficult cases.
Overall, these results support that our approach offers a better accuracy–efficiency trade-off than fixed pipelines, including prior affordance agents such as A4-Agent.

\begin{table}[!t]
    \centering
    \caption{Efficiency comparison on ReasonAff and UMD under a unified evaluation setup. Results are reported as mean $\pm$ standard deviation over three runs. ``Det.\,\&\,Seg.'' is our minimal baseline that invokes detection and segmentation once per sample, ``Full Fixed Chain'' executes all available skills in a pre-defined order, and A4-Agent~\cite{zhang2025a4agent} is included as a closely related agentic affordance baseline.}
    \vspace{-0.5em}
    \label{tab:efficiency}
    \adjustbox{max width=\textwidth}{
    \begin{tabular}{l|ccccc|ccccc}
        \toprule
        \multirow{2}{*}{\textbf{Method}}
            & \multicolumn{5}{c|}{\textbf{ReasonAff}}
            & \multicolumn{5}{c}{\textbf{UMD}} \\
        \cmidrule(lr){2-6}\cmidrule(lr){7-11}
            & \textbf{gIoU}$\uparrow$
            & \textbf{cIoU}$\uparrow$
            & \textbf{Tokens}$\downarrow$
            & \textbf{Lat.(s)}$\downarrow$
            & \textbf{Cost(\$)}$\downarrow$
            & \textbf{gIoU}$\uparrow$
            & \textbf{cIoU}$\uparrow$
            & \textbf{Tokens}$\downarrow$
            & \textbf{Lat.(s)}$\downarrow$
            & \textbf{Cost(\$)}$\downarrow$ \\
        \midrule
        Det.\,\&\,Seg.
            & 51.86$_{\pm0.42}$ & 43.73$_{\pm0.37}$ & 3{,}842$_{\pm96}$ & 4.1$_{\pm0.2}$ & 1.92$_{\pm0.05}$
            & 46.53$_{\pm0.45}$ & 37.77$_{\pm0.41}$ & 3{,}710$_{\pm88}$ & 3.9$_{\pm0.2}$ & 1.86$_{\pm0.04}$ \\
        Full Fixed Chain
            & 55.05$_{\pm0.56}$ & 49.57$_{\pm0.49}$ & 9{,}578$_{\pm141}$ & 12.3$_{\pm0.4}$ & 4.79$_{\pm0.09}$
            & 50.19$_{\pm0.52}$ & 49.24$_{\pm0.46}$ & 9{,}215$_{\pm133}$ & 11.8$_{\pm0.3}$ & 4.61$_{\pm0.08}$ \\
        A4-Agent (GPT-4o) \cite{zhang2025a4agent}
            & 56.87$_{\pm0.61}$ & 52.15$_{\pm0.55}$ & 6{,}978$_{\pm124}$ & 10.1$_{\pm0.6}$ & 3.88$_{\pm0.11}$
            & 50.92$_{\pm0.58}$ & 49.77$_{\pm0.51}$ & 6{,}747$_{\pm117}$ & 9.6$_{\pm0.4}$ & 3.74$_{\pm0.06}$ \\
        \midrule
        ours
            & \textbf{60.53$_{\pm0.48}$} & \textbf{54.91$_{\pm0.44}$} & \textbf{5{,}247$_{\pm109}$} & \textbf{5.8$_{\pm0.2}$} & \textbf{2.62$_{\pm0.05}$}
            & \textbf{52.74$_{\pm0.47}$} & \textbf{50.04$_{\pm0.43}$} & \textbf{5{,}831$_{\pm114}$} & \textbf{6.4$_{\pm0.2}$} & \textbf{2.92$_{\pm0.05}$} \\
        \bottomrule
    \end{tabular}
    }
\end{table}

\subsubsection{Sensitivity analysis on $C_{tt}$.}
\begin{figure}[!t]
    \centering
    \includegraphics[width=\textwidth]{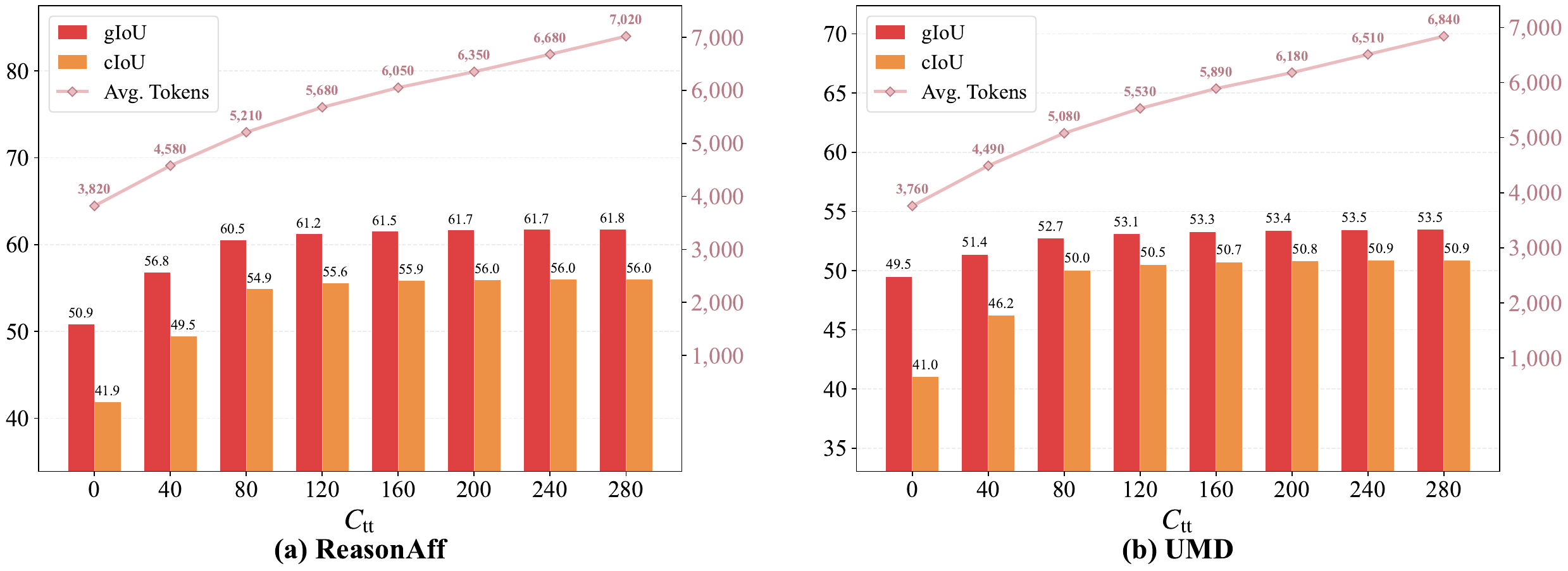}
    \vspace{-1.5em}
    \caption{Sensitivity of episodic memory capacity $C_{\textsc{tt}}$ on (a)~ReasonAff and (b)~UMD.
    Bars show gIoU and cIoU in \% (left axis); the line shows average input-token consumption per sample (right axis).
    All runs use GPT-4o as the decision brain.}
    \label{fig:ctt_sensitivity}
\end{figure}

As illustrated in Fig.~\ref{fig:ctt_sensitivity}, while increasing the episodic bank capacity yields monotonic improvements in gIoU and cIoU across both ReasonAff and UMD, it simultaneously incurs a proportional escalation in average token consumption per sample. This performance gain eventually reaches a saturation point, where the benefit of broader experience coverage is offset by distractive semantic interference from contextually divergent yet semantically similar episodes. Such a plateau suggests that beyond an optimal threshold, additional memory provides diminishing returns while inflating the computational footprint. Consequently, selecting a balanced $C_{tt}$ is paramount to achieving superior grounding accuracy while preserving an efficient accuracy-cost Pareto frontier.

\subsubsection{Sensitivity analysis on Commitment Gating Thresholds.}
\begin{figure}[!t]
    \centering
    \includegraphics[width=\textwidth]{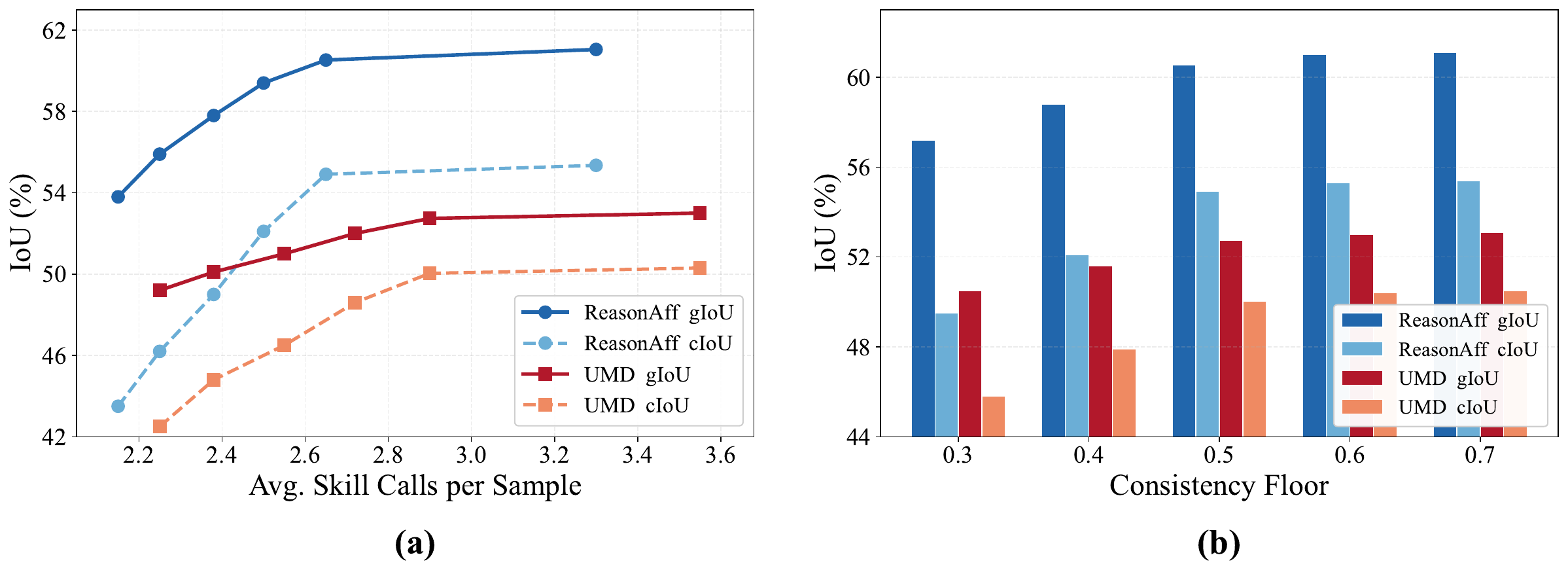}
    \vspace{-1.5em}
    \caption{Sensitivity analysis of commitment gating thresholds on ReasonAff and UMD.
    \textbf{(a)}~gIoU and cIoU vs.\ average skill calls per sample under varying $\delta \in \{0.4, 0.5, 0.6, 0.7, 0.8, 0.9\}$ (left to right along each curve), with $\underline{\omega}$ fixed at 0.5.
    Each marker corresponds to one $\delta$ setting; the circled marker denotes the default $\delta{=}0.8$.
    \textbf{(b)}~gIoU and cIoU under varying consistency floor $\underline{\omega} \in \{0.3, 0.4, 0.5, 0.6, 0.7\}$, with $\delta$ fixed at 0.8.}
    \label{fig:gating_sensitivity}
\end{figure}

As shown in Fig.~\ref{fig:gating_sensitivity}, both $\delta$ and $\underline{\omega}$ exhibit a consistent accuracy--cost trade-off.
A lower $\delta$ yields lenient gating with fewer retries but weaker grounding quality; raising $\delta$ enforces stricter evidence acceptance, triggering targeted retries for ambiguous cases and steadily improving gIoU and cIoU---yet at the expense of more average skill calls.
Beyond $\delta{=}0.8$, performance saturates while inference cost continues to grow, indicating diminishing returns.
The consistency floor $\underline{\omega}$ follows the same pattern: higher values prevent premature commitment under tool disagreement, but the benefit plateaus beyond $\underline{\omega}{=}0.5$.
These results confirm that our default configuration ($\delta{=}0.8$, $\underline{\omega}{=}0.5$) sits at the knee of the accuracy--cost Pareto frontier.

\subsubsection{Sensitivity analysis on Retrieval Count $N$.}
Table~\ref{tab:topn_ablation} reports the effect of varying the number of retrieved memory references $N$ fed to the Router.
Performance rises sharply from $N{=}1$ to $N{=}2$, with the majority of the accuracy gain already captured by just two retrieved references, and a marginal further improvement at $N{=}3$.
Beyond that, grounding quality plateaus and slightly declines at $N{\ge}5$, as semantically adjacent but contextually divergent entries introduce distracting routing priors.
These results indicate that a small number of retrieved references suffices to seed the Router with effective priors. We therefore use $N{=}2$ as the default setting for the main experiments, since it captures most of the benefit with lower retrieval overhead, while $N{=}3$ is reported as the best-performing setting in this ablation.

\begin{table}[!t]
    \centering
    \caption{Sensitivity of retrieval count $N$ on ReasonAff and UMD.}
    \vspace{-0.5em}
    \label{tab:topn_ablation}
    \adjustbox{max width=\textwidth}{
    \begin{tabular}{c|cccc|cccc}
        \toprule
        \multirow{2}{*}{$N$}
            & \multicolumn{4}{c|}{\textbf{ReasonAff}}
            & \multicolumn{4}{c}{\textbf{UMD}} \\
        \cmidrule(lr){2-5}\cmidrule(lr){6-9}
            & \textbf{gIoU}$\uparrow$ & \textbf{cIoU}$\uparrow$
            & \textbf{$P_{50}$}$\uparrow$ & \textbf{$P_{50\text{-}95}$}$\uparrow$
            & \textbf{gIoU}$\uparrow$ & \textbf{cIoU}$\uparrow$
            & \textbf{$P_{50}$}$\uparrow$ & \textbf{$P_{50\text{-}95}$}$\uparrow$ \\
        \midrule
        1  & 57.34 & 50.12 & 61.85 & 40.22 & 49.81 & 46.53 & 53.10 & 26.34 \\
        2  & 60.53 & 54.91 & 66.73 & 45.53 & 52.74 & 50.04 & 57.62 & 29.85 \\
        3  &  \textbf{60.89} & \textbf{55.11} & \textbf{67.26} & \textbf{45.67} & \textbf{53.01} & \textbf{50.44} & \textbf{57.84} & \textbf{29.93} \\
        5  & 60.31 & 54.40 & 66.41 & 45.20 & 52.48 & 49.80 & 57.30 & 29.61 \\
        8  & 59.87 & 53.92 & 65.78 & 44.53 & 52.10 & 49.41 & 56.82 & 29.12 \\
        10 & 59.42 & 53.35 & 65.10 & 43.98 & 51.83 & 48.96 & 56.35 & 28.70 \\
        \bottomrule
    \end{tabular}
    }
\end{table}

\subsubsection{Sensitivity analysis on Verifier Weights $\alpha$, $\beta$, $\gamma$}
The Verifier aggregates three diagnostic signals via $v_t = \alpha\,\omega_t + \beta\,\zeta_t + \gamma\,\sigma(\mu_t)$, where $\alpha$, $\beta$, $\gamma$ weight cross-skill consistency, cross-scale stability, and evidence sufficiency, respectively.
Table~\ref{tab:weight_ablation} evaluates eight configurations with $\alpha+\beta+\gamma=1$.
Ablating any single signal consistently degrades performance across both datasets, with consistency ($\alpha$) exerting the largest influence---corroborating its role as a hard commitment floor.
Stability ($\beta$) contributes more to localization precision (cIoU, $P_{50}$) while sufficiency ($\gamma$) provides a secondary safety net against premature commitment.
Over-emphasizing any single dimension also hurts performance, confirming that balanced weighting is preferred.
The highlighted row achieves the best overall accuracy--cost trade-off and is adopted as the default configuration.

\begin{table}[!t]
    \centering
    \caption{Ablation of Verifier weights $(\alpha, \beta, \gamma)$ on ReasonAff and UMD ($\alpha+\beta+\gamma=1$).
    The highlighted row denotes the configuration used in our main experiments.}
    \vspace{-0.5em}
    \label{tab:weight_ablation}
    \adjustbox{max width=\textwidth}{
    \begin{tabular}{l|ccc|cccc|cccc}
        \toprule
        \multirow{2}{*}{\textbf{Configuration}}
            & \multirow{2}{*}{$\alpha$} & \multirow{2}{*}{$\beta$} & \multirow{2}{*}{$\gamma$}
            & \multicolumn{4}{c|}{\textbf{ReasonAff}}
            & \multicolumn{4}{c}{\textbf{UMD}} \\
        \cmidrule(lr){5-8}\cmidrule(lr){9-12}
            & & &
            & \textbf{gIoU}$\uparrow$ & \textbf{cIoU}$\uparrow$ & \textbf{$P_{50}$}$\uparrow$ & \textbf{$P_{50\text{-}95}$}$\uparrow$
            & \textbf{gIoU}$\uparrow$ & \textbf{cIoU}$\uparrow$ & \textbf{$P_{50}$}$\uparrow$ & \textbf{$P_{50\text{-}95}$}$\uparrow$ \\
        \midrule
        Uniform           & 0.33 & 0.33 & 0.33 & 57.82 & 51.10 & 63.45 & 42.31 & 50.31 & 47.22 & 54.88 & 27.43 \\
        w/o $\omega$      & 0.00 & 0.60 & 0.40 & 53.14 & 45.78 & 58.32 & 37.90 & 48.62 & 44.05 & 52.10 & 25.18 \\
        w/o $\zeta$       & 0.60 & 0.00 & 0.40 & 57.40 & 50.93 & 63.02 & 42.05 & 49.88 & 46.51 & 54.42 & 27.08 \\
        w/o $\mu$         & 0.60 & 0.40 & 0.00 & 56.85 & 49.62 & 62.41 & 41.58 & 49.20 & 45.83 & 53.74 & 26.72 \\
        $\omega$-heavy    & 0.60 & 0.25 & 0.15 & 59.10 & 53.22 & 65.18 & 44.20 & 51.85 & 48.30 & 56.45 & 28.92 \\
        $\zeta$-heavy     & 0.25 & 0.60 & 0.15 & 55.34 & 49.01 & 60.75 & 40.33 & 51.20 & 47.75 & 55.88 & 28.14 \\
        $\mu$-heavy       & 0.25 & 0.25 & 0.50 & 56.73 & 50.18 & 62.10 & 41.82 & 50.45 & 46.98 & 55.02 & 27.61 \\
        \textbf{Default (Ours)} & \textbf{0.50} & \textbf{0.30} & \textbf{0.20}
            & \textbf{60.53} & \textbf{54.91} & \textbf{66.73} & \textbf{45.53}
            & \textbf{52.74} & \textbf{50.04} & \textbf{57.62} & \textbf{29.85} \\
        \bottomrule
    \end{tabular}
    }
\end{table}
\subsection{Case study.}
\label{app:case_study}
\begin{figure}[!tp]
    \centering
    \includegraphics[width=\textwidth]{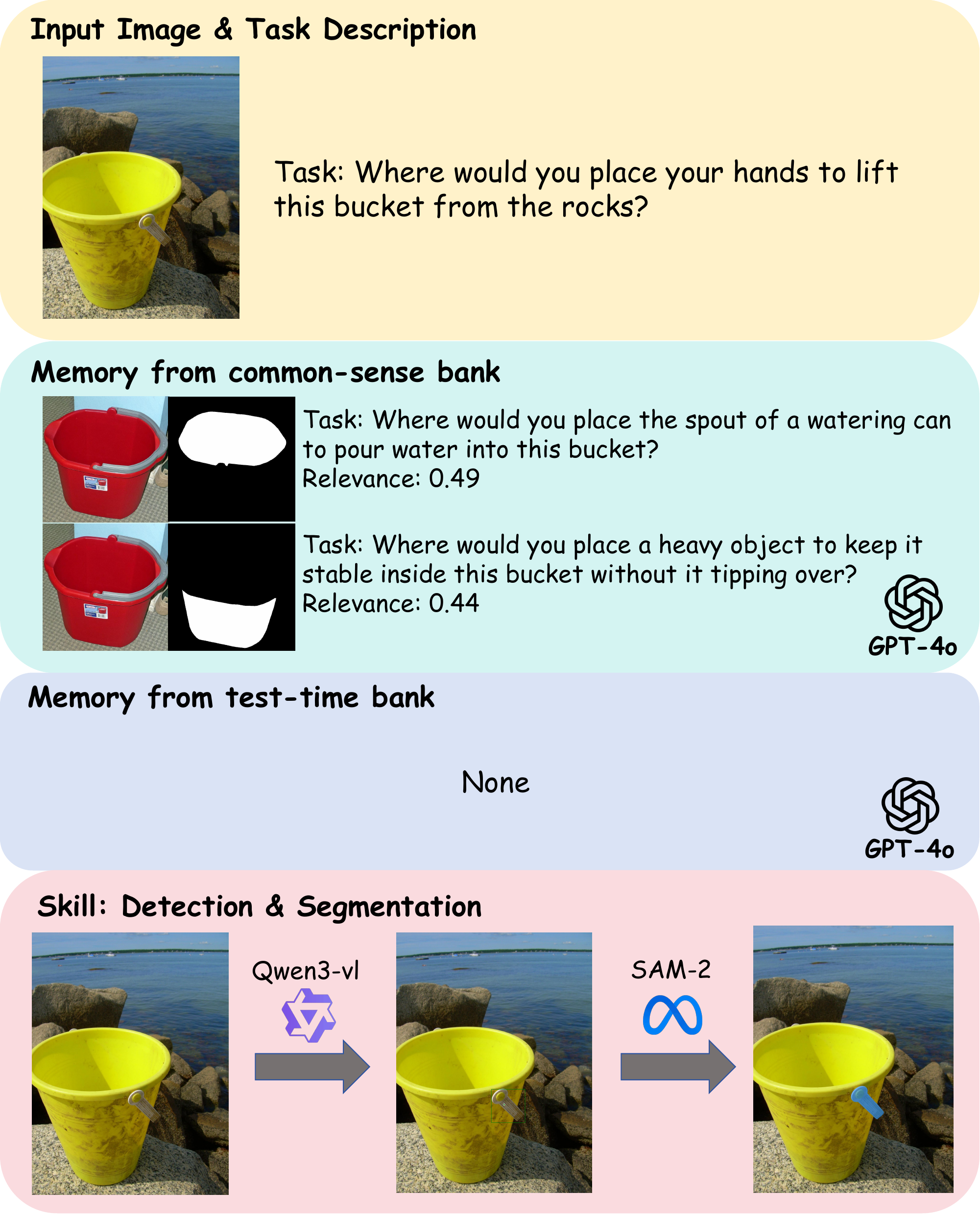}
    \vspace{-1.5em}
    \caption{The first page of the intermediate results of case 1.}
    \label{fig:case1-1}
\end{figure}

\begin{figure}[!tp]
    \centering
    \includegraphics[width=\textwidth]{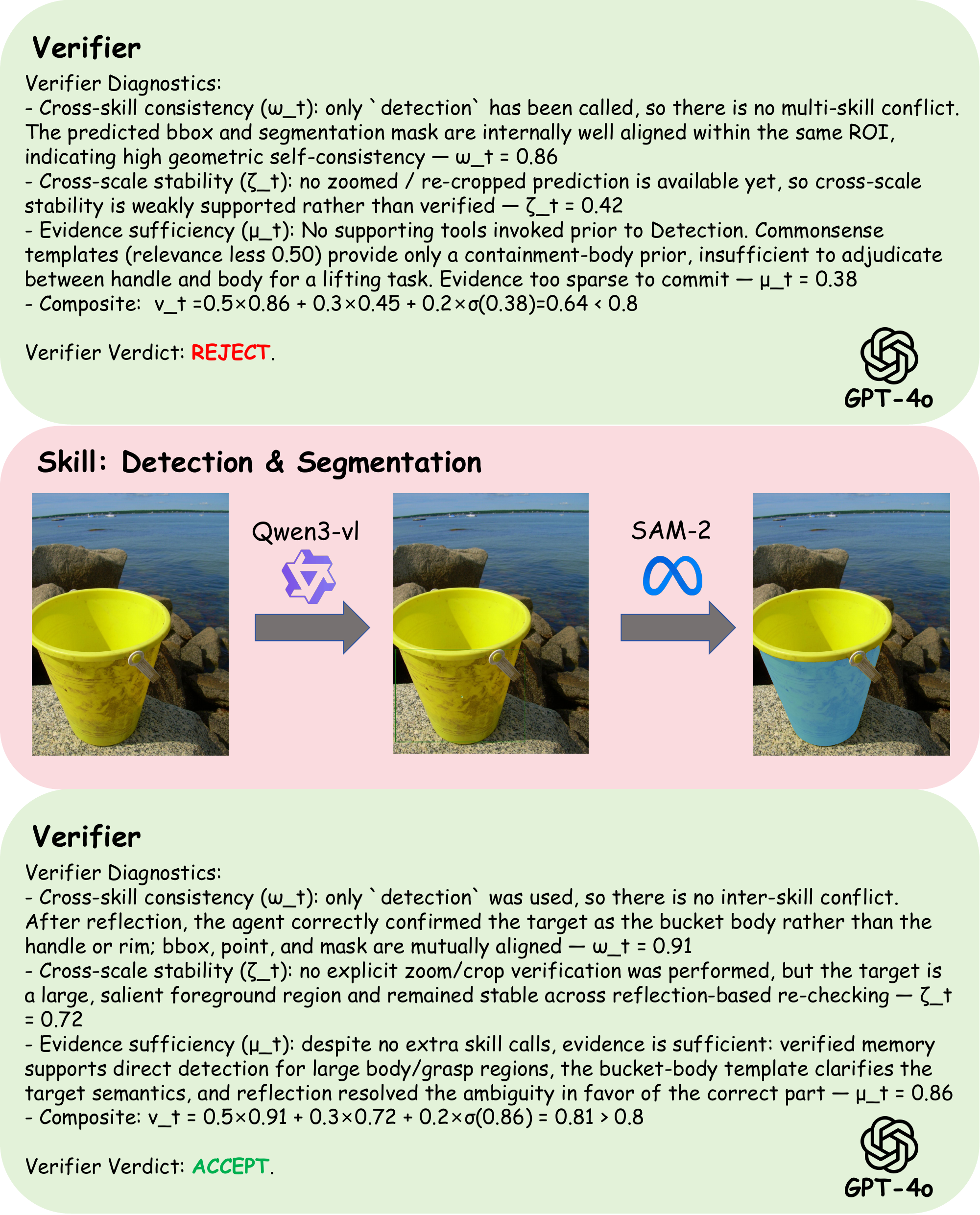}
    \vspace{-1.5em}
    \caption{The second page of the intermediate results of case 1.}
    \label{fig:case1-2}
\end{figure}

\begin{figure}[!tp]
    \centering
    \includegraphics[width=\textwidth]{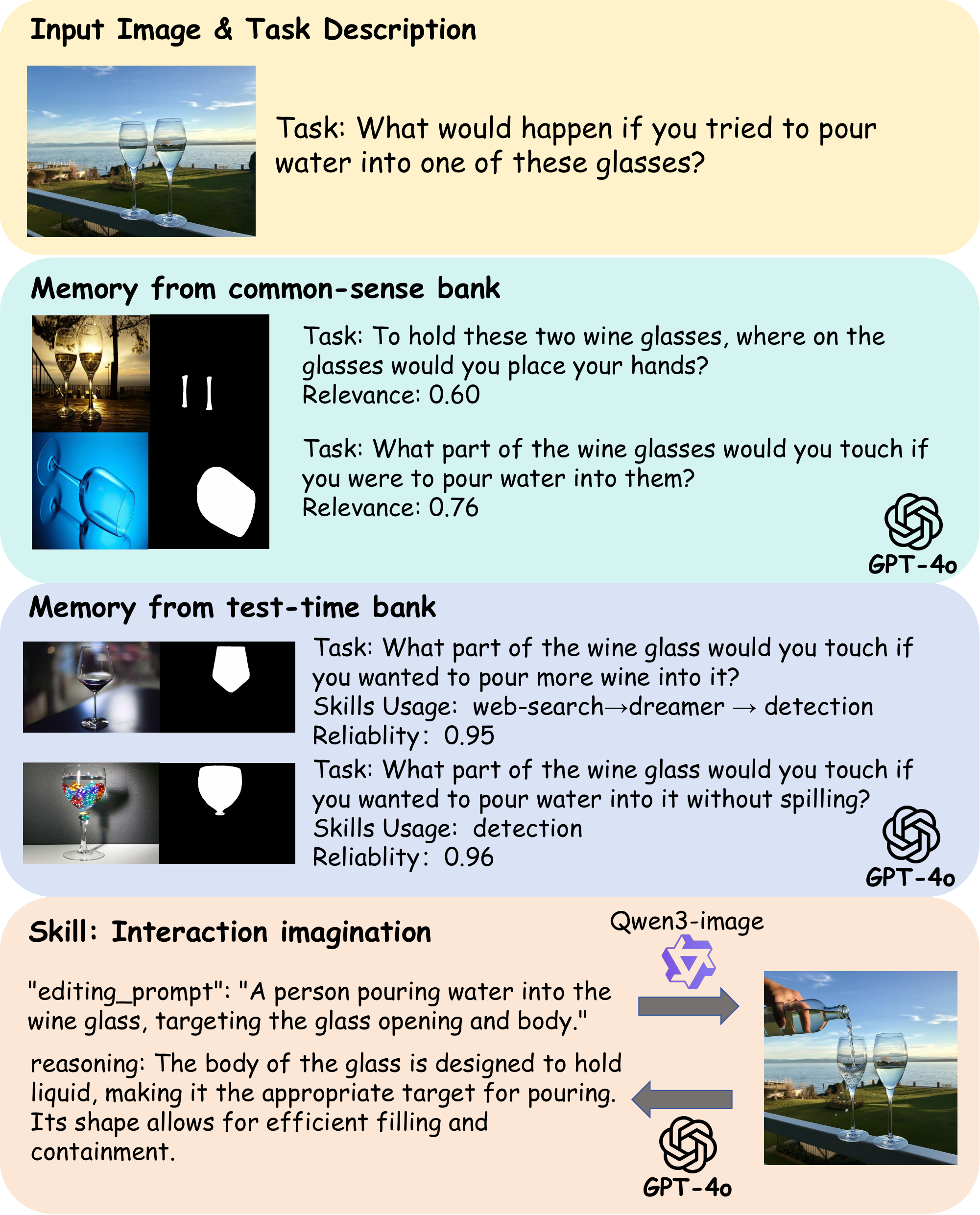}
    \vspace{-1.5em}
    \caption{The first page of the intermediate results of case 2.}
    \label{fig:case2-1}
\end{figure}

\begin{figure}[!tp]
    \centering
    \includegraphics[width=\textwidth]{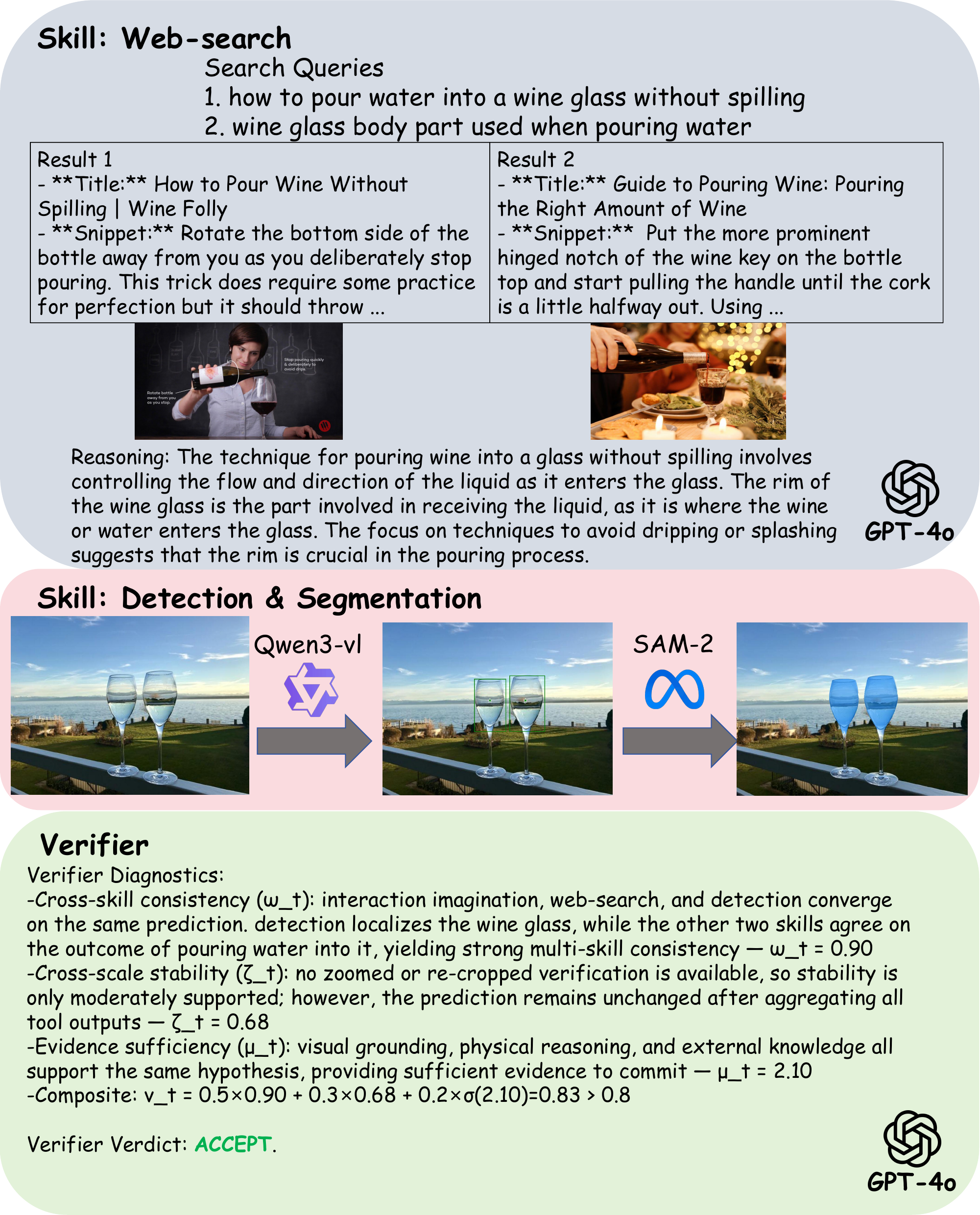}
    \vspace{-1.5em}
    \caption{The second page of the intermediate results of case 2.}
    \label{fig:case2-2}
\end{figure}

\begin{figure}[!tp]
    \centering
    \includegraphics[width=\textwidth]{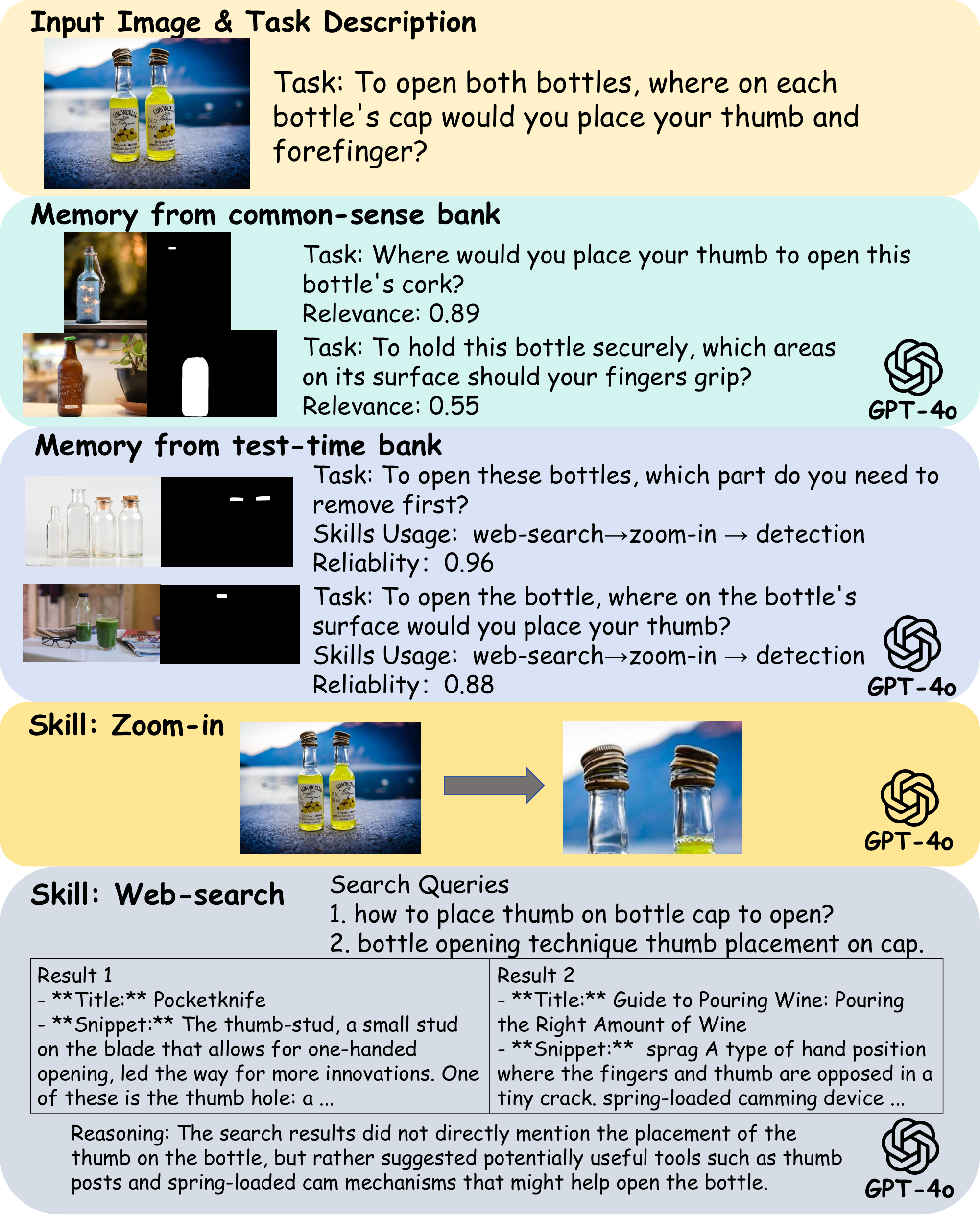}
    \vspace{-1.5em}
    \caption{The first page of the intermediate results of case 3.}
    \label{fig:case3-1}
\end{figure}

\begin{figure}[!tp]
    \centering
    \includegraphics[width=\textwidth]{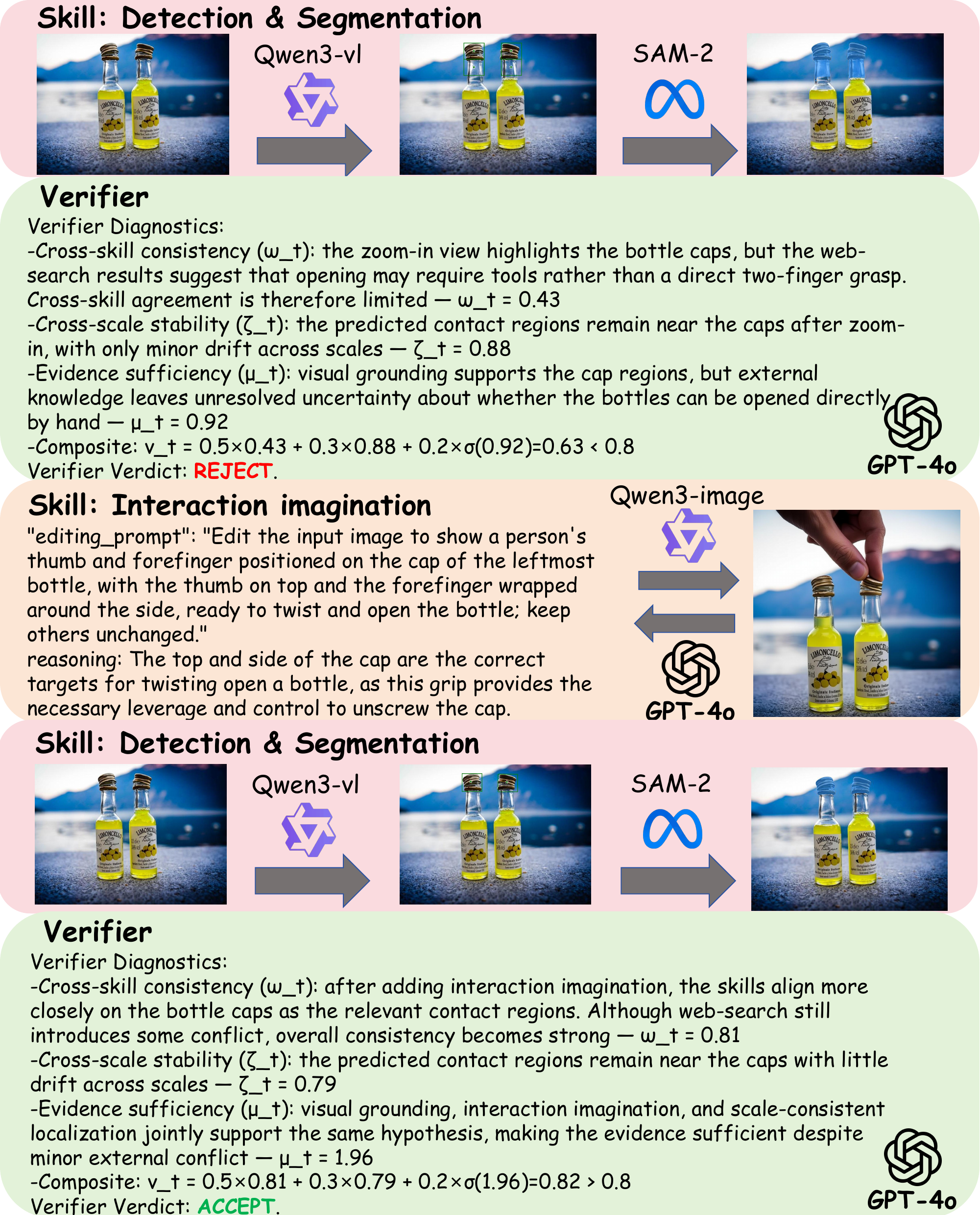}
    \vspace{-1.5em}
    \caption{The second page of the intermediate results of case 3.}
    \label{fig:case3-2}
\end{figure}

We provide three representative case studies that expose typical failure modes in affordance grounding (Figs.~\ref{fig:case1-1}--\ref{fig:case3-2}). Across the examples, the useful behavior is not simply that more tools are called, but that retries are targeted: when evidence is inconsistent, incomplete, or unstable, the Verifier denies commitment and the Router selects the skill most relevant to the diagnosed gap. The resulting predictions rely on cross-tool agreement, cross-scale stability, and evidence sufficiency rather than any single intermediate output. Compared with fixed-workflow affordance agents such as A4-Agent~\cite{zhang2025a4agent}, this design makes the execution path depend on the observed failure mode, which is the key mechanism behind the qualitative improvements.

\section{Limitations and Future Work}
\label{app:limitations}
Our work has several limitations. First, current affordance datasets still underrepresent long-tail object categories, unusual designs, and interaction patterns that arise in real environments; this limits how thoroughly memory construction and verifier thresholds can be stress-tested. Second, the system does not yet learn from large-scale interaction feedback. Its routing and verification policies rely on pretrained reasoning priors plus hand-designed diagnostics, which may be insufficient for harder out-of-distribution cases where repeated physical feedback would be informative. Future work should expand affordance data coverage and study reinforcement learning or other feedback-driven training methods for improving recovery, planning, and adaptation in embodied settings.